\documentclass[sn-mathphys,Numbered]{sn-jnl}

\usepackage{natbib}
\usepackage{graphicx}%
\usepackage{multirow}%
\usepackage{amsmath,amssymb,amsfonts}%
\usepackage{amsthm}%
\usepackage{mathrsfs}%
\usepackage[title]{appendix}%
\usepackage{xcolor}%
\usepackage{textcomp}%
\usepackage{manyfoot}%
\usepackage{booktabs}%
\usepackage{algorithm}%
\usepackage{algorithmicx}%
\usepackage{algpseudocode}%
\usepackage{listings}%

\usepackage{float}
\usepackage{pdfpages}
\usepackage{stmaryrd}
\usepackage{tikz}
\usepackage{pgfplots}
\usepackage{graphicx}
\usepackage{pgfplots,adjustbox}
\usepackage{lineno,hyperref,multicol,amsmath,longtable,enumerate}
\usepackage{scalefnt,ae,graphicx,epsfig,amsmath,amssymb,subfig,pifont,longtable}
\usepackage{filecontents}
\pgfplotsset{compat=1.12}
\usepackage{lscape}
\usepackage{orcidlink}
\usepackage{subfig}
\makeatletter
\newbox\sf@box
\newenvironment{SubFloat}[2][]%
{\def\sf@one{#1}%
	\def\sf@two{#2}%
	\setbox\sf@box\hbox
	\bgroup}%
{ \egroup
	\ifx\@empty\sf@two\@empty\relax
	\def\sf@two{\@empty}
	\fi
	\ifx\@empty\sf@one\@empty\relax
	\subfloat[\sf@two]{\box\sf@box}%
	\else
	\subfloat[\sf@one][\sf@two]{\box\sf@box}%
	\fi}
\makeatother

\theoremstyle{thmstyleone}%
\theoremstyle{thmstyletwo}%

\theoremstyle{thmstylethree}%

\begin{document}

\title[Article Title]{High-level hybridization of heuristics and metaheuristics to solve symmetric TSP: a comparative study}


\author*[1,2]{\fnm{Carlos Alberto} \sur{da Silva Junior}\orcidlink{0000-0003-4173-7087}}\email{carlosdamat@ufsj.edu.br}

\author[3]{\fnm{Roberto Yuji} \sur{Tanaka}\orcidlink{0000-0003-2496-2565}}\email{robertoyuji@gmail.com}

\author[3]{\fnm{Luiz Carlos Farias} \sur{da Silva}}\email{fariaslcfs@gmail.com}

\author[2,3]{\fnm{Angelo} \sur{Passaro}\orcidlink{0000-0002-2421-0657}}\email{angelopassaro@gmail.com}

\affil*[1]{\orgdiv{Departamento de Matem\'atica e Estat\'istica}, \orgname{Universidade Federal de S\~ao Jo\~ao del-Rei}, \orgaddress{\street{P\c{c} Frei Orlando}, \city{S\~ao Jo\~ao del-Rei}, \postcode{36.307-352}, \state{MG}, \country{Brazil}}}

\affil[2]{\orgdiv{PG-CTE}, \orgname{Instituto Tecnol\'ogico de Aeron\'autica - ITA}, \orgaddress{\street{P\c{c} Marechal Eduardo Gomes}, \city{S\~ao Jos\'e dos Campos}, \postcode{12.228-900}, \state{SP}, \country{Brazil}}}

\affil[3]{\orgdiv{Divis\~ao de F\'isica Aplicada}, \orgname{Instituto de Estudos Avan\c{c}ados - IEAv}, \orgaddress{\street{Trevo Coronel Aviador Jos\'e Alberto Albano do Amarante}, \city{S\~ao Jos\'e dos Campos}, \postcode{12.228-001}, \state{SP}, \country{Brazil}}}


\abstract{The Travelling Salesman Problem - TSP is one of the most explored problems in the scientific literature to solve real problems regarding the economy, transportation, and logistics, to cite a few cases. Adapting TSP to solve different problems has originated several variants of the optimization problem with more complex objectives and different restrictions. Metaheuristics have been used to solve the problem in polynomial time. Several studies have tried hybridising metaheuristics with specialised heuristics to improve the quality of the solutions. However, we have found no study to evaluate whether the searching mechanism of a particular metaheuristic is more adequate for exploring hybridization. This paper focuses on the solution of the classical TSP using high-level hybridisations, experimenting with eight metaheuristics and heuristics derived from k-OPT, SISR, and segment intersection search, resulting in twenty-four combinations. Some combinations allow more than one set of searching parameters. Problems with 50 to 280 cities are solved. Parameter tuning of the metaheuristics is not carried out, exploiting the different searching patterns of the eight metaheuristics instead. The solutions' quality is compared to those presented in the literature.}




\maketitle

\section{Introduction}\label{sec1}
\par The Travelling Salesman Problem (TSP) has been mentioned in salesman manuals since the early 19th century. Due to its practical nature and social interest, TSP is one of the most studied NP-hard problems. 
Solving TSP allows finding the shortest route for a journey from a list of cities whose distances between them are known. According to the model, each city can only be visited once, which leads to the development of complex combinatorial optimizations. In addition to the conventional heuristics developed to solve problems such as TSP, stochastic methods have been widely used in the search for solutions to problems of combinatorial nature, with an emphasis on metaheuristics \cite{2015Saji}.

\par Several combinatorial problems are inspired by the classic TSP, such as ATSP (Asymmetric TSP) \cite{1982Frieze}, CBTSP (Coloured Balanced Travelling  Salesman Problem) \cite{2015Li}, CTSP (Cluster Travelling  Salesman Problem) \cite{1975Chisman}, the GTSP (Generalized TSP) \cite{1983Laporte}, the PTSP (Probabilistic Travelling  Salesman Problem) \cite{1995Powell}, the PVDTSP (Polygon-Visiting Dubins Travelling  Salesman Problem) \cite{2012Obermeyer}, the TSP-D (Travelling  Salesman Problem with Drone) \cite{2021Roberti}; the TSPPD-H (Travelling  Salesman Problem with Pickups, Deliveries and Handling Costs) \cite{2010Battarra}, the TSPTW (Travelling  Salesman Problem with Time Window) \cite{1985Savelsbergh}, Time-Critical Maritime UAV Mission Planning \cite{2022DeLimaFilho}, VRP (Vehicle Routing Problems) \cite{1985Christofides}, to present a shortlist. In each of these new models, specific constraints are added to adjust the model to real situations.

\par Numerous exact or direct methods are presented in the literature to solve the TSP, such as those based on Branch and Bound \cite{1954Dantzig}, the Lagrangian Relaxation \cite{1970Held}, the k-OPT heuristic \cite{1973Lin}, Recurrent Neural Networks \cite{1985Hopfield}, and more recently, MILP (Compact Mixed-Integer Linear Program) \cite{2021Roberti} and the exact algorithm based on a branch-and-bound presented in Arigliano et al. \cite{2019Arigliano}.

\par TSP is a discrete problem with a finite space solution. However, as the number of cities increases, the number of possible solutions increases exponentially. A problem with n cities has $n!$ possible solutions. For a relatively small problem with $n = 100$ cities, more than $9x10^{157}$ candidate solutions are possible. This number grows to $7.9x10^{374}$ when a list of $200$ cities is considered. The use of direct methods to obtain solutions to problems with a large number of cities becomes unfeasible due to the computational time required to obtain the optimal solution.

\par Stochastic techniques to solve discrete problems has been a natural way out. In this context, metaheuristics have been widely applied in diverse studies obtaining good solutions, not necessarily the optimal one, in a reasonable time, especially in instances with a large number of cities, or nodes, and the most varied types of restrictions. 

\par Real-world problems based on TSP, such as logistics and transport, has been solved satisfactorily in an acceptable time in the last years \cite{2019Dong} introduced a variation of the genetic algorithm, named NGA (Novel Genetic Algorithm), to solve CBTSP and CTSP. But the authors pointed out that even with NGA obtaining a better response than the GA, it would be necessary to apply it in larger instances and other types of problems. Thus, studies of new generation algorithms to solve the CBTSP problem are pertinent.

\par Osaba et al. \cite{2018Osaba} solve the OAG-TSP (Open-Path and Asymmetric Green Travelling Salesman Problem), which includes minimising CO2 emission by a car in an urban environment. The authors have used three classic metaheuristics: SA (Simulated Annealing), TB (Tabu Search) and VNS (Variable Neighbourhood Search), in three different scenarios, each one with twenty nodes. However, the authors point out the need to impose additional restrictions to solve larger problems. They also suggest using more metaheuristics in the search for solutions.

\par Therefore, even with the relative success of metaheuristics in solving TSP-based problems in a reasonable time, the increase in the number of cities poses difficulties in obtaining good solutions that approach the optimal one, which led to using of hybrid techniques to improve the performance in finding good solutions in reduced time. Several high-level hybridization techniques are proposed in literature combining algorithms \cite{2017Ezugwu} or modifying move operators \cite{2018Hussain}. Low-level hybridizations are also the focus of active research, for instance, mixing operators from two or more algorithms \cite{2018Zhong}.

\par Ezugwu et al. \cite{2017Ezugwu} presented a hybrid SOS-SA (Simulated-Annealing based Symbiotic Organisms Search Optimization) algorithm to solve the TSP. The authors claim this is a high-level hybridization, as the SOS and SA algorithms are used interchangeably. The SA algorithm is used to update each individual of the ecosystem generated in the SOS phases. The authors highlighted the need to improve algorithm stability by studying its behaviour in larger problems.

\par Zhong et al. \cite{2017Ezugwu} introduce a PSO-based low-level hybridization for discrete problems. The velocity equation depends on a learning probability, and basic operators are redesigned to produce a new position, allowing the particle to learn both from the personal best of each particle and the features of the problem. To enhance its ability to escape premature convergence, the particle uses the Metropolis acceptance criterion of the SA algorithm to decide whether to accept a new solution. Four experiments were performed with problems of 318, 575, 1291 and 3038 nodes to compare different acceptance criteria and another 9 experiments with problems from 198 to 1577 nodes to compare hybridizations of CLPSO (Comprehensive Learning PSO) \cite{2006Liang} proposals: D-CLPSO (discrete CLPSO) and S-CLPSO (set-based CLPSO), based on D-PSO (discrete PSO) and S-PSO (Set-PSO) \cite{2010Chen}.

\par Osaba et al. \cite{2020Osaba} presents a survey of recent advances in the study of TSP. The authors present a list of articles that solve the TSP or any of its variations using metaheuristics. Additionally, the authors propose hybridizations of three bio-inspired metaheuristics, the BA (Bat Algorithms), FA (Firefly Algorithms) and PSO (Particle Swarm Optimization) algorithms with the NS (Novelty Search) \cite{2008Lehman} engine. Each of these algorithms and their hybridizations were applied to problems ranging from 30 to 124 nodes. The authors state that it is necessary to explore alternative methods and/or hybridizations to solve the TSP and its variations and, in addition, studies involving problems with many nodes.

\par Wu and Gao \cite{2017Wu} present a hybrid algorithm of the SA with a greedy heuristic to solve the classic TSP. Five problems were solved in the experiments, with the number of cities ranging from 30 to 1577 with percentage errors ranging from 2.31\% to 862\%. The authors highlight the need to improve the proposed hybrid algorithm, as it presented difficulties in finding good quality solutions for problems with many cities.

\par Eskandari et al. \cite{2019Eskandari} presents a hybrid algorithm of PSO, ACO and the k-opt heuristic to solve the classic TSP. Six experiments were performed with numbers of nodes between 29 and 100 and obtained errors ranging from 0.0\% to 2.67\%. Wang and Xu \cite{2017Wang} present a hybrid algorithm between PSO and the 4 Vertices and 3 Edges heuristic to search for optimal solutions for the classical TSP. PSO is used to find an initial path that is improved with heuristics. Experiments were carried out with 29 instances with numbers of cities varying between 51 and 575 and the errors obtained varying between 0.0\% and 3.92\%.

\par However, we have found a few studies to evaluate whether the searching mechanism of a particular metaheuristic is more adequate for exploring hybridization in solving TSP-based problems. Moreover, the metaheuristics seem to be chosen by the authors' preferences.

\par In general, using metaheuristics in solving discrete problems requires specialised operators or hybridizations. One of the problems in using metaheuristics to solve discrete problems as TSP is the guarantee of generating feasible solutions. The literature presents several methods for generating an initial route, such as the Nearest Neighbour Heuristic \cite{1956Flood}, by permutation \cite{2018Hussain,2017Wu}, by using metaheuristics \cite{2016Mazidi}, Multiple Phase Neighbourhood Search - GRASP for VRP \cite{2010Marinakis}, among others.  Tentative solutions provided by general metaheuristics are usually unfeasible because the graph generated can present repeated and absent cities. To create new candidate solutions from the previous one, it is usual to replace the general move operators with permutation operators. Alternatively, correcting heuristics are necessary to randomly include the absent cities in the positions of the graph occupied by repeated nodes, making the solution feasible.

\par In this work, high-level hybridization, which requires modifications of neither the metaheuristics nor the heuristics used in conjunction with the metaheuristics, is used to solve the classic symmetric TSP. Three basic heuristics are exploited: one based on SISR (Slack Induction by String Removals) \cite{2020Christiaens}, one based on k-OPT, and an uncrossing heuristic of segments. The algorithms are applied to solve problems with different numbers of cities. The hybridizations and results obtained are detailed in the next sections. 

\section{Basic Algorithms}\label{sec2}

\par The ``classical'' TSP optimization can be formulated as a closed graph problem, where nodes correspond to cities, and the edges connecting these nodes correspond to paths between the cities. An instance is composed of a set of $n$ nodes, where the distance between these nodes is known. Thus, solving a TSP problem is finding the shortest path, or route, to go through all the nodes, visiting each city only once, and with the start and end nodes coincident. Considering $c_{ij}$ as the cost of going from node $i$ to node $j$ and, $x_{ij}$ the edge connecting these nodes, we have that a formulation for the TSP can be given as:
\begin{equation}\label{eq:TSP}
	\begin{array}{cl} Minimize & f(X) = \displaystyle\sum_{i=1}^{n}\sum_{j=1}^{n}c_{ij}x_{ij} \\ Subject \ to: & \left\{\begin{array}{cc} \displaystyle\sum_{i=1}^{n}x_{ij} = 1, & \forall \ j \in \{1,2,\cdots,n\} \\ \displaystyle\sum_{j=1}^{n}x_{ij} = 1, & \forall \ i \in \{1,2,\cdots,n\} \\ x_{ij} \in \{0,1\} &  \forall \ i,j \in \{1,2,\cdots,n\} \end{array}\right. \end{array}.
\end{equation}

\par There are deterministic methods which allow finding the optimal solution, but the computational cost increases exponentially as the problem size increases. One possible option studied in the literature is stochastic algorithms. This text uses some metaheuristics coupled with heuristics adapted to solve the TSP optimization.

\subsection{Metaheuristics}\label{subsec1}

\par Metaheuristics are stochastic algorithms whose search patterns to find the optimum are inspired by biological, physical, social, ethnological behaviours, and mathematical concepts. Metaheuristics can be classified as single-solution-based (S metaheuristics) and population-based metaheuristics (P metaheuristics). S metaheuristics are memory-oriented algorithms in which each iteration improves a single solution, favouring a local search. On the other hand, P metaheuristics favour the broad exploration of the search space, as they use characteristics of a set of individuals at each iteration to improve the search process.

\par Many metaheuristics are proposed in Literature. For example, Evolutionary Algorithms (EA) are inspired by the theory of natural selection of species proposed by Darwin \cite{2009Talbi}. Particle Swarm Optimization (PSO) is a P-metaheuristic based on the social behaviour of birds \cite{1995Kennedy}. PSO uses personal best and best position of a group of individuals to update the position of each individual. Sine Cosine Algorithm (SCA) is a P-metaheuristic algorithm that explores trigonometry properties \cite{2016Mirjalili}. Vortex Search (VS) is an S-metaheuristic inspired on vortex pattern created by the vortical flow of the stirred fluids \cite{2015Dogan} and Modified Vortex Search (MVS) is a P-metaheuristics based in VS \cite{2016Dogan}. The Gravitational Search Algorithm (GSA) is a P-metaheuristic based on the law of gravity and mass interactions \cite{2009Rashedi}. The best solution has the largest mass in the GSA proposal. The Black Hole Algorithm (BH) is a P-metaheuristic inspired by the strong gravitational attraction of collapsed massive stars, the black holes. The black hole is the best solution found so far. A Schwarzschild radius, associated with the black hole, defines a region in which any tentative solution can be destroyed, generating a new tentative in any place of the search space, a kind of solution restart \cite{2015Kumar}. Simulated Annealing (SA) is an S-metaheuristic proposed in analogy to the annealing procedure of metalworking \cite{1983Kirkpatrick}. There are variants of each cited metaheuristic. For instance, Genetic Algorithms are a kind of Evolutionary Algorithm that usually deal with binary variables and use crossover operators. Still, evolutionary algorithms are also used with floating point or integer variables and without crossover operators. There are several variants of PSO, but, despite the inspiring metaphor, they differ in the choice of the search control parameters of the movement operators. 

\par The cited algorithms are used in this paper to solve TSP optimization. The algorithms are well documented in the referenced literature and are not reproduced here. As metaheuristics are stochastic algorithms, the tentative solutions usually present repeated cities and a lack of cities. To guarantee feasible solutions, a simple correction strategy can be adopted after the new set of tentative solutions is generated by the metaheuristics. The algorithm is depicted in Algorithm \ref{<algAlgorithm1>}.
\begin{algorithm}
	\caption{Tentative solution correction for feasibility guarantee.}\label{<algAlgorithm1>}
	\begin{enumerate}
		\item The metaheuristic generates a set of tentative solutions;
		\item Remove duplicate nodes of the tentative solutions; the first occurrence of a repeated node is not removed and remains in the original position;
		\item Create vector A containing the missing nodes;
		\item Complete the tentative solution chosen nodes randomly from vector A; the tentative solution becomes feasible, i.e., each city will be visited only once;
		\item Compute objective functions;
		\item If iterative process is not finished, Return to 1.
	\end{enumerate}
\end{algorithm}

\par The objective functions are evaluated from the corrected tentative solutions. The corrected tentative solution has to be returned to the metaheuristics to continue the search procedure. The cited correction does not avoid path crossing. Therefore, algorithms to address this issue are necessary to guarantee good solutions.

\subsection{SISR-based heuristic}\label{subsec2}
\par Schirimpf et al. \cite{2000Schrimpf} presents the idea of ``Ruin and Reconstruction'' (R\&R) as a ``all-purpose-heuristic'' and solve VRP problems. The authors ruin the solution based on a radial removal of nodes in the graph and suggest another type of removal, the random one, which removes nodes from the sequence based on a certain probability. According to the authors, there is a ``whole universe'' of methods to recreate the complete route. They adopt the ``best insertion'' heuristic for the reconstruction phase; that is, each node that was not yet part of the graph after the ruin phase is reinserted in the best position into the graph being reconstructed.

\par The SISR heuristic (Slack Induction by String Removals) was used in a hybrid way with the SA metaheuristics in solving problems based on VRP \cite{2020Christiaens}: Capacitated Vehicle Routing Problem (CVRP), Vehicle Routing Problem with Time Windows (VRPTW), and Pickup-and-Delivery Problem with Time-Windows (PDPTW). This heuristic is based on the R\&R principle; firstly, a process of destroying the routes is applied, with a mechanism to exclude graph edges. Then, a reconstruction phase is executed, for instance, by a greedy process. Christiaens and Berghe (2020) apply the SISR to solve the CVRP considering an unlimited fleet of vehicles located in the same depot and instances whose size ranges from 100 to 1000 customers. In this sense, optimising the route for each vehicle is desired, obeying that each customer needs to be visited only once by a single vehicle and that the carrying capacity of the vehicles is not violated. Therefore, the Classic TSP can be treated as a CVRP. The fleet is formed by a single vehicle with infinite capacity, and each city is a customer whose demand becomes irrelevant.

\par In this sense, a simplified heuristic based on SISR can be used for TSP optimization. The SISR-based heuristic used in this work can be described as follows: consider the graph of nodes generated by any metaheuristic with $n$ cities. Since design variables can take on any value between $1$ and $n$, the graph can have nodes with repeated values. The duplicate values are eliminated from the tentative solution, keeping only the first one that appears in the TSP graph. Thus, the places where duplicated nodes appear remain empty until the route is reconstructed. A vector $A$ is created to store all nodes of the set $\{1, 2, \cdots, n\}$ that is lacking in the TSP graph.

\par To finish the destruction phase based on the proposal of Christiaens and Berghe (2020), a node Seed is chosen randomly, and several consecutive edges are eliminated before and after the Seed. This number of edges to be eliminated is also chosen at random, ranging from $1$ to $\dfrac{n}{2}$. All the nodes that formed the consecutive edges that were eliminated in this phase are included in the vector $A$, thus forming the set with all the nodes that need to be reallocated in the graph. Therefore, at the end of the destruction phase, there is a graph with some nodes associated with their initial positions and a set $A$ with all the nodes of the graph that still do not have an associated position.

\par To improve the performance of hybridization with SISR, the graph can be divided into equal parts, and consequently, each of these parts is a branch of connected edges. Each of these branches is called a subtour. The SISR destruction phase is applied within each subtour: a Seed node is chosen within each subtour, some edges are deleted, and the nodes deleted within each subtour are sent to vector $A$. The reconstruction phase is based on the principle of better inclusion. First, for each node $n_i$ in the graph, a vector $V_i$ is created, called the neighbourhood of node $n_i$, where all other nodes are inserted in increasing order of distance from node $n_i$. Then, for each position $j$ of the graph with no associated node, it is searched in the neighbourhood $V_{j-1}$ of the node $n_{j-1}$, the node still left in the set $A$ with the smallest distance from node $n_{j-1}$ to insert it at position $j$. At the end of this process, a feasible and updated graph is returned to the metaheuristics evolution process. It is important to note that the SISR-based heuristic is applied to each solution generated by the metaheuristics.

\par Algorithm \ref{<algAlgorithm2>} extends Algorithm \ref{<algAlgorithm1>} and is summarized in the following.
\begin{algorithm}
	\caption{SISR for TSP.}\label{<algAlgorithm2>}
	\begin{enumerate}
		\item The metaheuristic generates a set of tentative solutions;
		\item Remove duplicate nodes of the tentative solution; the first occurrence of a repeated node is not removed and remains in the original position;
		\item Create vector A containing the missing nodes;
		\item For a preset number of subtours, do:
		\begin{enumerate}
			\item Selects and remove an additional node of the tentative solution at random, including it in vector A;
			\item Removes at random a continuous sequence of edges around the node chosen in step 4.(a).;
			\item Includes the nodes related to the deleted edges in vector A;
		\end{enumerate}
		\item Rebuild the tentative solution: for each vacancy in the tentative solution, selects in A the node whose position is closest to the node that precedes the vacancy and includes it in the solution vector, removing it from A. The process continues until emptying A, and results in a tentative solution in which each city will be visited only once;
		\item Compute objective functions.
	\end{enumerate}
\end{algorithm}

\par Three aspects of the SISR-based heuristic used here are noteworthy: first, all the solutions generated by the metaheuristics are made feasible, the procedure to generate new solutions provided by the metaheuristics during the optimization process is maintained, and finally, the node neighbourhood information is used to accelerate the convergence process.

\subsection{3-OPT based heuristic}\label{subsec3}
\par The k-OPT heuristics \cite{1973Lin} consist of inspecting changes in the way of reconnecting k non-consecutive edges of the graph to reduce the path length. The 2k node vertices can be reconnected in several ways, and the combination that generates the shortest length is chosen, reorganizing the sequence of nodes of the tentative solution. The process is repeated until no new set of k deleted edges improves the length value of the route. Increasing the value of k, the number of path length evaluations grows quadratically since the number of possible recombination's of the extremes of the k excluded non-consecutive edges is $\dfrac{2k(2k-3)}{2}$.

\par In this work, the heuristic based on 3-OPT is used to verify if its hybridization with the metaheuristics helps find good solutions for the TSP. Three non-consecutive edges are chosen at random, say $\overline{P_{i-1}P_i}$, $\overline{P_{j-1}P_j}$ and $\overline{P_{l-1}P_l}$, with $i < j,l$,, and search for the smallest subgraph $P_{i-1} - P_{m_{1}} - P_{m_{2}} - P_{m_{3}} - P_{m_{4}} -  P_{m_{5}} - P_{i-1}$, with $m_1, \cdots, m_5 \in {i,j-1,j,l-1,l}$. Finally, the points $P_{m_{1}}, P_{m_{2}}, P_{m_{3}}, P_{m_{4}}$, and $P_{m_{5}}$ are positioned at the respective positions $i,j-1,j,l-1,l$, of the graph. This process based on the 3-OPT heuristic is repeated, in each solution, a certain number of times that may or may not be related to the number of nodes in the graph. Algorithm \ref{<algAlgorithm3>} illustrates the procedure.
\begin{algorithm}
	\caption{3-OPT based heuristic.}\label{<algAlgorithm3>}
	Given a feasible solution (See algorithm \ref{<algAlgorithm1>}):
	\begin{enumerate}
		\item Repeat a preset number of times:
		\begin{enumerate}
			\item Select three non-consecutive edges of the tentative solution;
			\item Removes the nodes of the selected edges from the tentative solution, storing them in vector A;
			\item Select, keeping the first element fixed, the combination of nodes in A with the shortest length;
			\item The tentative solution is filled in again, placing the nodes in vector A in the vacant positions in the sequence of the shortest route obtained;
		\end{enumerate}
		\item The obtained solution is feasible and returns to the metaheuristic.
	\end{enumerate}
\end{algorithm}

\subsection{The Uncross heuristic}\label{subsec4}
\par The uncrossing heuristic searches in the final routes for possible ``crossings'' between edges. The edges uncrossing heuristic traverse the final solution returned by the metaheuristics searching for crossings, undoing them by changing the position of a limited number of nodes in the tentative solution and keeping the other nodes in their positions.

\par Consider that the edges $\overline{P_{i-1}P_i}$ and $\overline{P_{j-1}P_j}$, with $i<j$, intersect. Consider $P_{i-1} = (x_1,y_1)$, $P_{i} = (x_2,y_2)$, $P_{j-1} = (x_3,y_3)$, and $P_{j} = (x_4,y_4)$. Thus, the lines containing the edges  $\overline{P_{i-1}P_i}$ and $\overline{P_{j-1}P_j}$ are given by, respectively, $$r_1: (x,y) = (x_1 + s(x_2-x_1), y_1 - s(y_2-y_1)), s \in \mathbb{R}, \text{ and }$$  $$r_2: (x,y) = (x_3 + t(x_4-x_3), y_3 - t(y_4-y_3)), t \in \mathbb{R}.$$

\par Thus, the lines $r_1$ and $r_2$ intersect if the linear system k formed by them in the variables s and t has a unique solution:
\begin{equation}\label{eq:SistemaIntesecaoSegmentos}
	k: \left\{\begin{array}{ccccc} s(x_2-x_1) & - & t(x_4-x_3) & = & x_3-x_1 \\ s(y_2-y_1) & - & t(x_4-x_3) & = & x_3-x_1 \end{array}\right..
\end{equation}
where $s,t \in [0,1]$.

\par
\begin{figure}[H]
	\centering
	\begin{tikzpicture}[scale=0.75,transform shape]
		\draw[red] (2,-3) -- (1,-1.5);
		\fill[red] (1,-1.5) circle (1mm) node[below right,red] {$P_{i+1}$};
		\draw[red] (4,-1) -- (1,-1.5);
		\fill[white] (4,1) circle (1mm) node[above,red] {$...$};
		\draw[red] (-3,-1) -- (-1.5,0);
		\fill[white] (-5,0) circle (1mm) node[above,red] {$...$};
		\draw[red] (3,2) -- (4,1.3);
		\draw[red] (4,0.9) -- (4,-1);
		\fill[red] (4,-1) circle (1mm) node[right,red] {$P_{i+2}$};
		\draw[red] (-3,-1) -- (-5,0);
		\fill[red] (3,2) circle (1mm) node[above] {$P_{j-3}$};
		\draw[red] (2,1) -- (1.5,-0.5);
		\draw[red] (2,1) -- (3,2);
		\fill[red] (2,1) circle (1mm) node[above left] {$P_{j-2}$};
		\fill[red] (-3,-1) circle (1mm) node[left] {$P_{i-2}$};
		\fill[blue] (-1.5,0) circle (1mm) node[above] {$P_{i-1}$};
		\fill[blue] (2,-3) circle (1mm) node[right] {$P_{i}$};
		\fill[blue] (-3,-3) circle (1mm) node[above] {$P_{j}$};
		\fill[blue] (1.5,-0.5) circle (1mm) node[above right] {$P_{j-1}$};
		\fill[white] (-5,-2) circle (1mm) node[above,red] {$...$};
		\fill[red] (-4,-2) circle (1mm) node[above] {$P_{j+1}$};
		\draw[blue] (-1.5,0) -- (2,-3);
		\draw[red] (-3,-3) -- (-4,-2);
		\draw[red] (-5,-2) -- (-4,-2);
		\draw[blue] (-3,-3) -- (1.5,-0.5);
	\end{tikzpicture}
	\caption{Illustration of edges crossing present in a graph formed by the points of the branch $P_{i-1} - P_{i} - P_{i+1} - P_{i+2} - \cdots - P_{j-3} - P_{j-2} - P_{j-1} - P_{j}$.}
	\label{fig:ilustracaoDeLacoNoGrafo}
\end{figure}
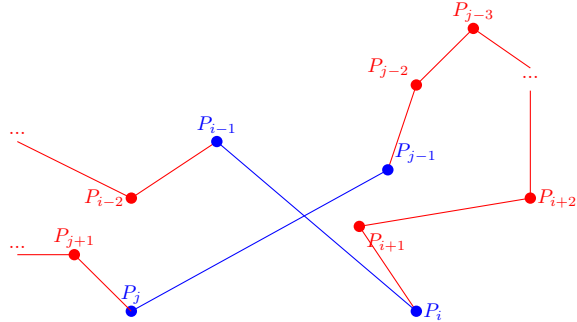

 Consider the edges crossing illustrated in Figure \ref{fig:ilustracaoDeLacoNoGrafo}, formed by the points of the branch $P_{i-1} - P_{i} - P_{i+1} - P_{i+2} - \cdots - P_{j-3} - P_{j-2} - P_{j-1} - P_{j}$.

\par The edges crossing provokes an inversion of the travel direction, for instance, from clockwise to anti-clockwise. 

\par
\begin{figure}[H]
	\centering
	\begin{tikzpicture}[scale=0.75,transform shape]
		\draw[red] (2,-3) -- (1,-1.5);
		\fill[red] (1,-1.5) circle (1mm) node[below right,red] {$P_{i+1}$};
		\draw[red] (4,-1) -- (1,-1.5);
		\fill[white] (4,1) circle (1mm) node[above,red] {$...$};
		\draw[red] (-3,-1) -- (-1.5,0);
		\fill[white] (-5,0) circle (1mm) node[above,red] {$...$};
		\draw[red] (3,2) -- (4,1.3);
		\draw[red] (4,0.9) -- (4,-1);
		\fill[red] (4,-1) circle (1mm) node[right,red] {$P_{i+2}$};
		\draw[red] (-3,-1) -- (-5,0);
		\fill[red] (3,2) circle (1mm) node[above] {$P_{j-3}$};
		\draw[red] (2,1) -- (1.5,-0.5);
		\draw[red] (2,1) -- (3,2);
		\fill[red] (2,1) circle (1mm) node[above left] {$P_{j-2}$};
		\fill[red] (-3,-1) circle (1mm) node[left] {$P_{i-2}$};
		\fill[blue] (-1.5,0) circle (1mm) node[above] {$P_{i-1}$};
		\fill[blue] (2,-3) circle (1mm) node[right] {$P_{i}$};
		\fill[blue] (-3,-3) circle (1mm) node[above] {$P_{j}$};
		\fill[blue] (1.5,-0.5) circle (1mm) node[above right] {$P_{j-1}$};
		\fill[white] (-5,-2) circle (1mm) node[above,red] {$...$};
		\fill[red] (-4,-2) circle (1mm) node[above] {$P_{j+1}$};
		\draw[blue,dashed] (-1.5,0) -- (2,-3);
		\draw[red] (-3,-3) -- (-4,-2);
		\draw[red] (-5,-2) -- (-4,-2);
		\draw[blue,dashed] (-3,-3) -- (1.5,-0.5);
		\draw[thick] (-1.5,0) -- (1.5,-0.5);
		\draw[thick] (-3,-3) -- (2,-3);
	\end{tikzpicture}
	\caption{Illustration of the correction of a part of the graph after the uncrossing operation.}
	\label{fig:ilustracaoDeLacoNoGrafo2}
\end{figure}
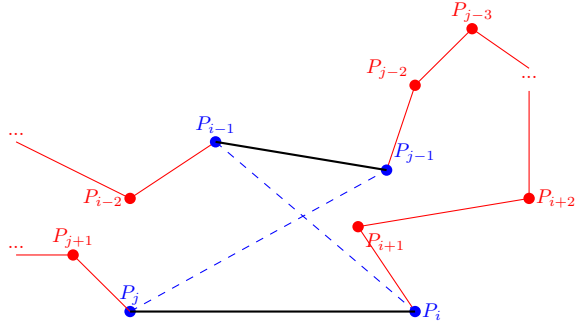

 Therefore, to undo the edges crossing, it is necessary to invert the nodes of the intersecting segments and, in addition, to invert the direction of travel on the branch of the graph between the nodes $P_{i-1}$ and $P_{j}$. Linking the nodes $P_{i-1}$ with $P_{j-1}$ and $P_{i}$ with $P_{j}$, as illustrated in Figure \ref{fig:ilustracaoDeLacoNoGrafo2}, the intersection is eliminated and, to maintain the direction of travel, the new sequence of nodes between $P_{i-1}$ with $P_{j}$, in the graph, is given by $$\cdots - P_{i-2} - P_{i-1} - {\color{blue} P_{j-1} - P_{j-2} - P_{j-3} - \cdots - P_{i+2} - P_{i+1} - P_{i}} - P_{j} - P_{j+1} - \cdots.$$

\par
\begin{figure}[H]
	\centering
	\begin{tikzpicture}[scale=0.75,transform shape]
		\draw[red] (1.5,-0.5) -- (1,-1.5);
		\fill[blue] (1,-1.5) circle (1mm) node[below right] {$P_{i+1}$};
		\draw[blue] (4,-1) -- (1,-1.5);
		\fill[white] (4,1) circle (1mm) node[above,red] {$...$};
		\draw[red] (-3,-1) -- (-1.5,0);
		\fill[white] (-5,0) circle (1mm) node[above,red] {$...$};
		\draw[red] (3,2) -- (4,1.3);
		\draw[red] (4,0.9) -- (4,-1);
		\fill[blue] (4,-1) circle (1mm) node[right] {$P_{i+2}$};
		\draw[red] (-3,-1) -- (-5,0);
		\fill[red] (3,2) circle (1mm) node[above] {$P_{j-3}$};
		\draw[blue] (2,1) -- (2,-3);
		\draw[red] (2,1) -- (3,2);
		\fill[blue] (2,1) circle (1mm) node[above left] {$P_{j-2}$};
		\fill[red] (-3,-1) circle (1mm) node[left] {$P_{i-2}$};
		\fill[red] (-1.5,0) circle (1mm) node[above] {$P_{i-1}$};
		\fill[blue] (2,-3) circle (1mm) node[right] {$P_{i}$};
		\fill[red] (-3,-3) circle (1mm) node[above] {$P_{j}$};
		\fill[red] (1.5,-0.5) circle (1mm) node[above left] {$P_{j-1}$};
		\fill[white] (-5,-2) circle (1mm) node[above,red] {$...$};
		\fill[red] (-4,-2) circle (1mm) node[above] {$P_{j+1}$};
		\draw[red] (-1.5,0) -- (1.5,-0.5);
		\draw[red] (-3,-3) -- (-4,-2);
		\draw[red] (-5,-2) -- (-4,-2);
		\draw[red] (-3,-3) -- (2,-3);
	\end{tikzpicture}
	\caption{Illustration of the ineffectiveness of the uncrossing operation changing only the nodes of the edges involved without inverting the inner points that generate the loop.}
	\label{fig:ilustracaoDeLacoNoGrafo3}
\end{figure}
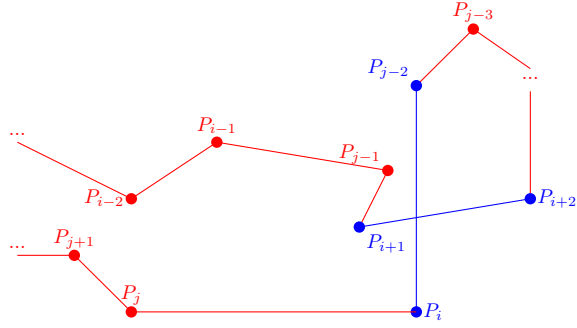

 The simple exchange of the points $P_{j-1}$  and $P_i$, without the inversion of the path in the branch, will not guarantee the elimination of crossings. In Figure \ref{fig:ilustracaoDeLacoNoGrafo3}, for example, the simple swap would lead to the branch $$P_{i-1} -{\color{blue} P_{j-1}} - P_{i+1} - P_{i+2} - \cdots - P_{j-3} - P_{j-2} - {\color{blue}P_{i}} - P_{j},$$ which would generate a new cross between the segments $\overline{P_{i+1}P_{i+2}}$ and $\overline{P_{j-2}P_{i}}$. Thus, to undo the edges crossing, the inversion of the sequence of points in the branch is essential.

\par
\begin{figure}[H]
	\begin{minipage}[b]{0.48\linewidth}
		\includegraphics[width=\linewidth]{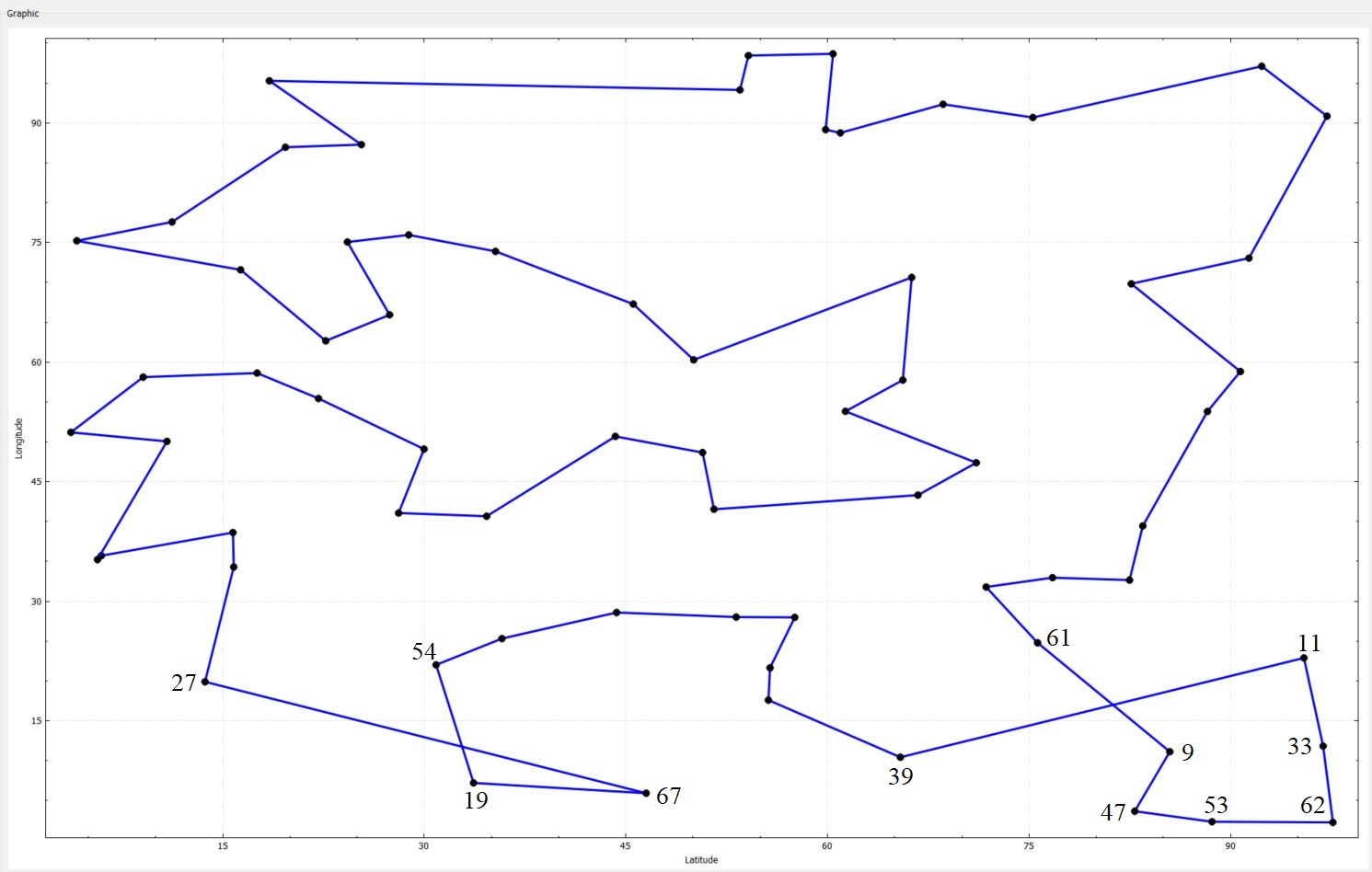}
		
		\caption{Final route for the Silalahi68 problem before correction by the uncrossing heuristic.}\label{fig4:ExplicacaoUncrossAC}
	\end{minipage}
	\hfill
	\begin{minipage}[b]{0.48\linewidth}
		\includegraphics[width=\linewidth]{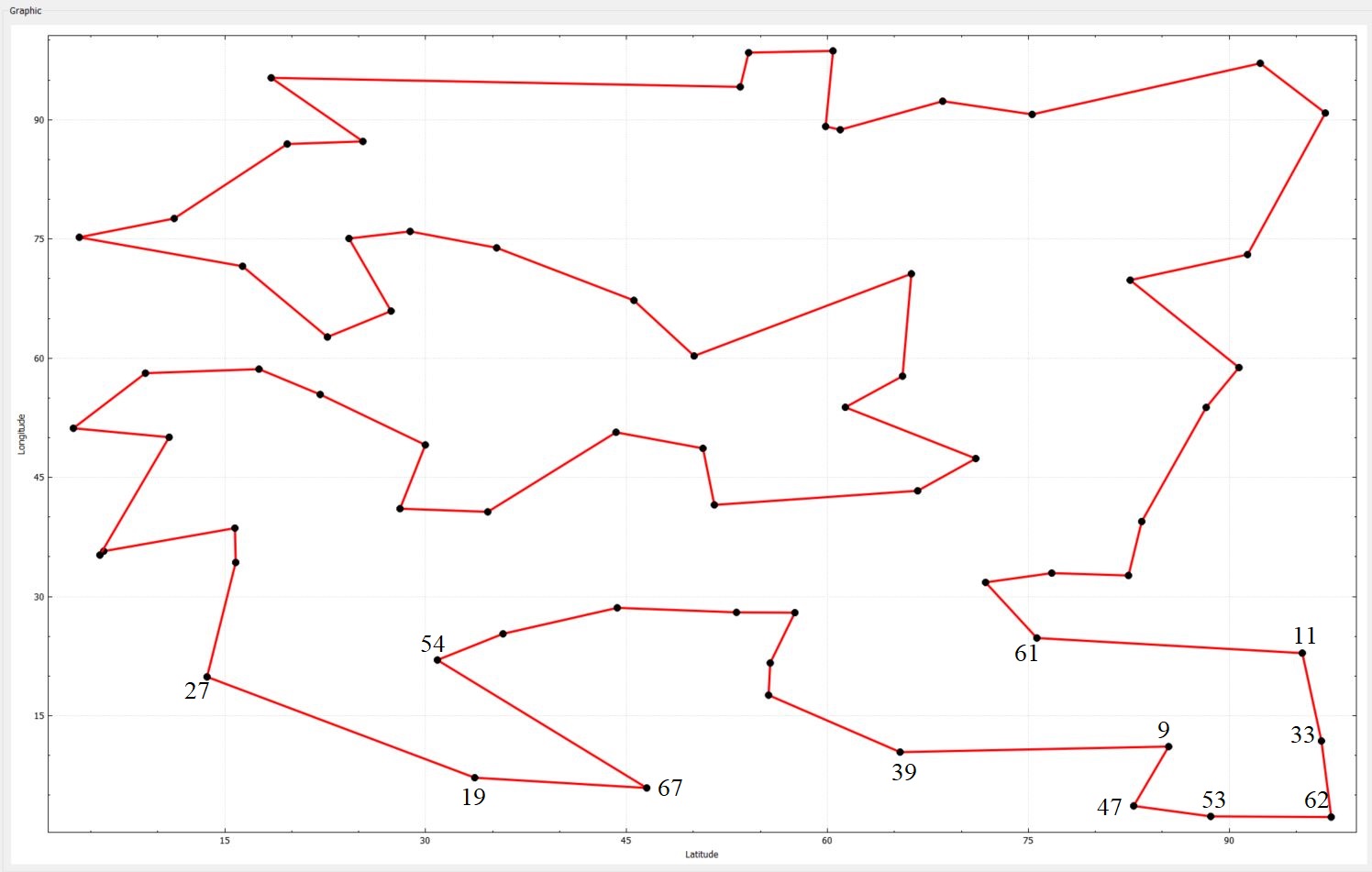}
		
		\caption{Final route for the Silalahi68 problem after the correction by the uncrossing heuristic.} \label{fig4:ExplicacaoUncrossDC}
	\end{minipage}
\end{figure}

 The exclusion of a crossing may reduce the traversed path. To illustrate the use of the segment uncrossing heuristic, observe the final route presented in Figures \ref{fig4:ExplicacaoUncrossAC} and \ref{fig4:ExplicacaoUncrossDC} obtained by one of the metaheuristics for the Silalahi68 problem \cite{2022Silalahi}. This route, before correction by the uncrossing heuristic, had a length of FO = 704. In the Figure \ref{fig4:ExplicacaoUncrossDC} the route after the correction is shown and the new value of the length of the route is now FO = 689, a reduction in the final route of 2.1\%.

\par Note that before the correction (Figure \ref{fig4:ExplicacaoUncrossAC}) there were two crossings. The first one, named $L1$, is formed by the branch of nodes $61-9-47-53-62-33-11-39$ and the second one, $L2$, is formed by nodes $54-19-67-27$. Thus, as in each branch the $P_{j-1}$ and $P_j$ (last node before the crossing and the first after, respectively) need to be maintained, it follows that the new branches to be included in the graph, replacing $L1$ and $L2$ are $M1: 61-11-33-62-53-47-9-39$ and $M2: 54-67-19-27$, respectively, and all other nodes of the graph are kept in their respective positions, generating the route shown in Figure \ref{fig4:ExplicacaoUncrossDC}.
\begin{algorithm}
	\caption{Uncross heuristic.}\label{<algAlgorithm4>}
	Given a feasible solution (See algorithm \ref{<algAlgorithm1>}):
	\begin{enumerate}
		\item For all $i,j \in \{1,2,\cdots,n\}$ and for each pair of edges $\overline{P_{i-1}P_i}$ and $\overline{P_{j-1}P_j}$, with $i<j$, do:
		\item if $\overline{P_{i-1}P_i}$ and $\overline{P_{j-1}P_j}$ generate a crossover:
		\begin{enumerate}
			\item undo the crossing by changing the sequence of points $P_{i-1} - P_{i} - P_{i+1} - P_{i+2} - \cdots - P_{j-3} - P_{j-2} - P_{j-1} - P_{j}$ the tentative solution by the sequence $P_{i-1} - P_{j-1} - P_{j-2} - \cdots - P_{i+2} - P_{i+1} - P_{i} - P_{j}$, so that the crossing is undone, the direction of travel of the nodes in the graph is restored and 
			\item increment i;
		\end{enumerate}
		\item else increment i;
		\item Repeat the process a preset number of times.
	\end{enumerate}
\end{algorithm}

\section{Methodology}\label{sec3}

\par This work concerns hybrid algorithms based on combinations of heuristics of solution reconstruction and classical general metaheuristics to solve Symmetrical TSP. Input data comprises the coordinates of each city and a neighbourhood matrix containing the distances between the cities. The solution is a vector containing all cities presented in the input data, with no repeated elements. 

\par The optimization problem intends to minimize the route length and crossing paths. Crossing paths are treated as restrictions. In the scope of this study, the visiting points have Euclidean coordinates (plane or quasi-plane problems). Problems with locations expressed in terms of geographical or spherical coordinates should consider the geodesic model, which is not taken into account. The metaheuristics EA, BH, GSA, MVS, PSO, SA, SCA and VS, are used. Each algorithm uses the same stopping criterion based on the number of objective function calls, and the parameters that define the metaheuristic search pattern are those suggested in the literature. A computational optimization framework developed in-house is used. All metaheuristics share the same control mechanism for high-level hybridization, to create and evaluate the tentative solutions, the parallelization procedure, and computational implementation.

\par Each metaheuristic randomly generates tentative solution vectors, resulting in repeated elements and a lack of visiting points. The moving operators of the metaheuristics are not specialized. In these cases, a correcting strategy is used, removing repeated nodes and including the absent ones, making the tentative solutions feasible (Algorithm \ref{<algAlgorithm1>}). We will refer to the solutions obtained with any metaheuristic and the correction strategy, with no hybridization, generically as TSP-MH.

\par The heuristics used in the hybridizations are summarized in Table \ref{tab1:Hibridizacoes} and were further described in the previous section.

\begin{table}[h]
	\centering
	\caption{Nomenclature used in each experiment to evaluate the high-level hybridizations between metaheuristics and elementary heuristics to solve the symmetric TSP.}
	\label{tab1:Hibridizacoes}
	\begin{tabular}{cl}
		\hline
		Name             &   Descriptions \\ \hline
		3OPTYRep         & Each tentative solution generated by the metaheuristic is corrected, becoming \\
		& feasible. In the following, the heuristic based on 3OPT (Algorithm \ref{<algAlgorithm3>})\\
		&  is applied. The process is repeated Y times. \\ \hline
		SISRXtours       & Each tentative solution is reconstructed and subdivided into X sub-tours, \\
		& and in each of these subtours, the SISR-based heuristic is applied \\
		& (Algorithm \ref{<algAlgorithm2>}). \\ \hline
		SISR3OPTYRep     & The SISR-based heuristic with 1 subtour is applied (Algorithm 2) in each \\
		& tentative solution. After reconstructing the route, \\
		& the heuristic based on 3OPT is repeated Y times (Algorithm 3). \\ \hline
		UNCROSSXProbYRep &  The SISR-based heuristic with 1 subtour is applied in each tentative solution. \\
		& After reconstructing the route, the segment uncrossing  heuristic is probability \\
		& applied on three straight edges, chosen randomly with an X exchange \\
		& rate. The uncrossing procedure is a special case of the general uncrossing \\
		& algorithm presented in Section \ref{subsec4}. The uncrossing is repeated Y times\\
		&  for each tentative solution in each iteration. \\ \hline
	\end{tabular}
\end{table}

\par At the end of the optimization process, a final loop of the uncrossing algorithm is applied on all final routes (Algorithm 4). The results obtained before and after the last uncrossing correction are compared. 

\par Two experiment sets are performed. In each experiment, different instances are solved by the hybrid algorithms. A group of benchmark problems from the TSPLIB95 \cite{1991Reinelt} library and from two other public data sets \cite{TSPCanada,TSPBurkardt} provides the instances used to evaluate the hybrid approaches proposed. 

\par The first experiment centres on solving a group of benchmark instances with the same number of nodes. The cases are from the TSPLIB95 \cite{1991Reinelt}. Fixing the problem size allows for illustrating each method's convergence and solutions quality, and it is the first step to identifying more promising metaheuristics to perform the second experiment set.

\par The second experiment set is composed of instances of different sizes, i.e., different number of nodes. Only a subset of metaheuristics used in the first experiment is explored in the second one. The main focus is the study of the hybrid methods as a function of the instance size, which intends to highlight the effect of different search mechanisms in exploring the solution space.

\par The population size of the population-based metaheuristics and the number of neighbours used for single-solution-based metaheuristics are numerically equal. In each experiment, a problem from the databases is solved at least thirty times, and the stopping criterium is the predefined number of objective function calls. The performance analysis is based on comparing the best solution found by the hybrid algorithms, Best, and the best solution registered either in the databases or a result presented in more recent publications, $Best_{ref}$. The solution error is calculated by:
\begin{equation} \label{eq:erro}
	Error (\%) = \left(\frac{Best - Best_{ref}}{Best_{ref}}\right) \times 100.
\end{equation}

\par The best solution obtained and the number of objective function calls required to get it are both used to identify the hybridization configurations that reach the best quality solutions.

\section{Results and Discussions}\label{sec4}
\par This section presents the results obtained for the two experiments, the first with three instances of 100 nodes and different optimal routes and the second with instances of different sizes.  

\par As expressed previously, we use eight metaheuristics, i.e., EA, BH, GSA, MVS, PSO, SA, SCA and VS combined at a high level with heuristics already presented in the literature. 

\par The control parameters of each metaheuristic used in the hybridizations are given in Table 2. The combination of heuristics are presented in Table 1, and the control parameters are presented in Table 3. 
\begin{table}[h]
	\caption{Parameters used by the metaheuristics.}
	\label{tab1:Metaheuristicas}
	\begin{tabular}{cl}
		\hline
		Metaheuristics &   Parameters \\ \hline
		BH-Black Hole  & No parameters \\ \hline
		EA-Evolutionary  & Mutation: 10 permutations \\ & No crossover \\ & Elitism: 1 \\ \hline
		GSA-Gravitational Search & Attractive individuals: 5 \\ & Initial Gravity: 100 \\ & Alpha: 20 \\ \hline
		MVS-Modified Vortex Search & The same as VS, but using ten vortex centres. \\ \hline
		PSO- Particle Swarm & Inertia Rate: 0.715 \cite{Jiang2007} \\ & Particle learning Rate: 1.7 \\ & Swarm learning rate: 1.7 \\ \hline
		SA-Simulated Annealing & Temperature decreasing: Hajek Prescription \cite{Laarhoven1987}, constant 0.9 \\ & Initial temperature: 2000 \\ & Final temperature: 1e-5  \\ \hline
		SCA- Sine Cosine & Maximum search radius: 2 \\ \hline
		VS- Vortex Search & No parameters. Neighbourhood decreases \\ & according to the Incomplete gamma function \\ \hline
	\end{tabular}
\end{table}
\begin{table}[h]
	\caption{Parameters used for each Heuristic. The percentage refers to the number of cities of each instance.}
	\label{tab1:HibridizacoesConfiguracoes}
	\begin{tabular}{cl}
		\hline
		Name &   Descriptions \\ \hline
		3OPTYRep  & Y = 50\%, 100\% and 200\%. \\ \hline
		SISRXtours & X=1, 2, 3 and 4 tours \\ \hline
		SISR3OPTYRep & Y corresponding to1\%, 5\%, and 10\% repetitions. \\ \hline
		UNCROSSXProbYRep &  X equal to 20\%, 50\%, and 100\% repetitions and \\ & Y equal to 1\%, 5\% and 10\% \\ \hline
	\end{tabular}
\end{table}

\par Each hybridization uses the same stopping criterion to solve a given problem: objective function calls. The number of objective function calls depends on the problem size. The population/neighbourhood size used is ten individuals chosen based on results obtained in preliminary evaluations considering time spent and the quality of the solutions found for several problems size. The algorithms solve each instance 30 times.

\par The error for each metaheuristic is computed from the best result obtained, as defined in Eq. \ref{eq:erro}. 

\subsection{The First Experiment set: KroX100 instances}
\par For the first experiment, the instances KroA100, KroB100 and KroC100 are used. These problems were chosen because they present the same number of cities and different optimal routes.

\par The final uncrossing loop was the first aspect evaluated. The heuristic is applied to all obtained solutions to undo possible edge crossings. The heuristic checks if an edge crosses any others following it. The number of times the uncrossing algorithm is applied traversing the tentative solution is defined in the input data. A preliminary study was done to verify the effectiveness and the number of times the heuristic has to be applied to reduce the crossings significantly.

\par Table \ref{tab1:tabelaLoops} illustrates the number of tentative solutions that reach the best result in each iteration of the final uncrossing loop for KroA100 and KroB100. KroC100 shows the same behaviour. For instance, 49\% of the thirty tentative solutions found by the SISR2tours heuristic are not altered after the first iteration, and this number is 100\% in the third iteration. In this case, the same is true for the three instances. On the other hand, only after three iterations are found solutions provided by the hybrid algorithm 3OPT50Rep that achieve their best result.
\begin{table}[h]
	\caption{Percentage of the solutions that reach the best objective function after each iteration of the final uncrossing loop: (a) KroA100, and (b) KroB100.}
	\label{tab1:tabelaLoops}
	\begin{tabular}{c}
		\includegraphics[width=1.0\linewidth]{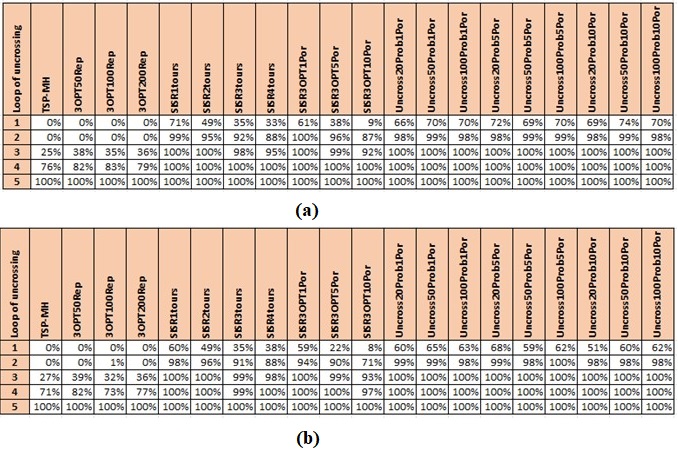}
	\end{tabular}
\end{table}

\par Notice that after three iterations, most of the hybrid algorithms' solutions have already reached the best possible solution. The exceptions, TSP-MH (metaheuristics with no hybridizations) and 3OPTYRep, start to be benefited in the third iteration. However, their final results are not good compared to the other hybrid algorithms tested.

\par
\begin{figure}[H]
	\includegraphics[width=\linewidth]{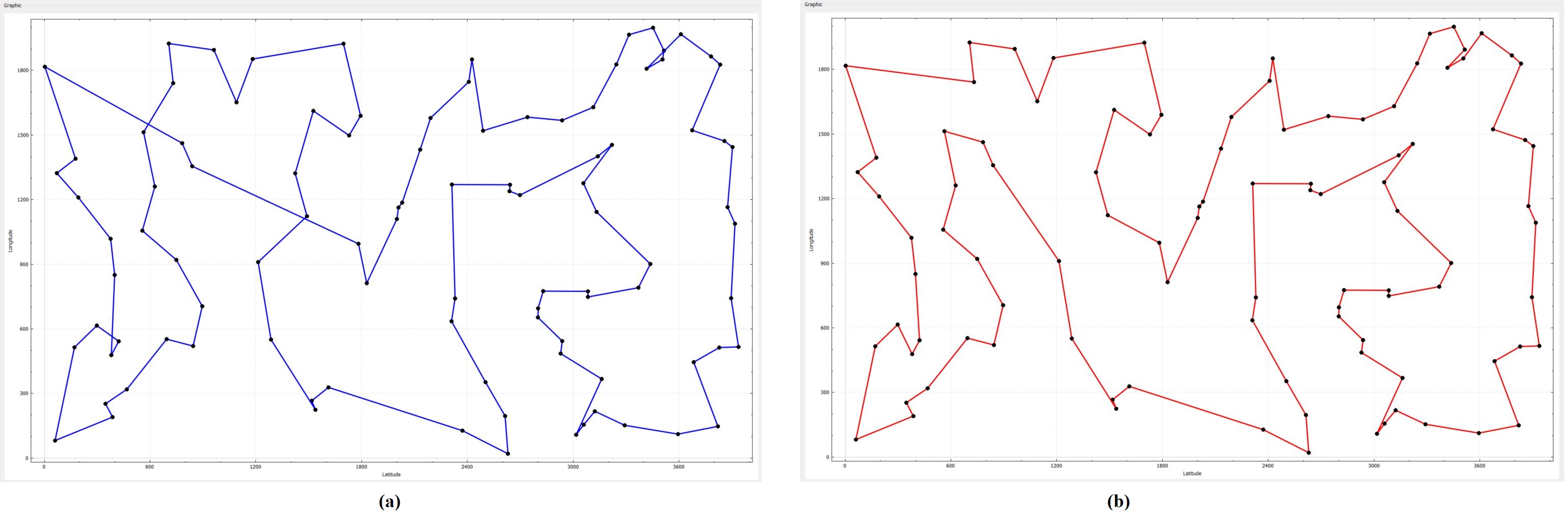}
	\caption{Best KroB100 route before correction (a) BU, with OF: 23341 and after correction (b) AU, with OF: 22647.}\label{fig5:ExplicacaoUncrossAC}
\end{figure}

 Applying the simple uncrossing heuristic at the end of the optimization allows for treating the imposed restrictions. For example, Figure \ref{fig5:ExplicacaoUncrossAC} shows one of the final routes obtained for the KroB100 problem, in the SISR4tours experiment, before correction - BU (Figure \ref{fig5:ExplicacaoUncrossAC}.a) and after correction - AU (Figure \ref{fig5:ExplicacaoUncrossAC}.b), respectively. Note that in the BU solution, there are four crosses, and the route length is 23341. After applying the uncrossing heuristic to correct the final route, the route length changed to 22647.

\par All results discussed from now on correspond to solutions after three iterations of the uncrossing algorithm.

\par The minimum e maximum errors related to the thirty runs of each of the three instances before (BU) and after (AU) the final uncrossing are presented in Table 5. TSP-MH, 3OPT50Rep, 3OPT100Rep, and 3OPT20Rep algorithms were not considered because they present the worst results, with errors far superior to the other options. This procedure has allowed a more objective comparison of the remaining hybridizations. For instance, their errors are all superior to 47\% after the final uncrossing loop.
\begin{table}[h]
	\centering
	\caption{Errors obtained before and after the final uncrossing algorithm.}
	\label{tab1:ErroACEDC}
	\begin{tabular}{ccccc}
		\hline
		Instance & Minimum & Maximum & Minimum & Maximum  \\
		& BU (\%) & BU (\%) & AU (\%) & AU (\%) \\ \hline
		KroA100 & 4.17 & 32.60 & 3.25 & 18.27 \\ \hline
		KroB100 & 4.29 & 31.24 & 2.29 & 12.76 \\ \hline
		KroC100 & 4.74 & 26.03 & 1.91 & 13.58 \\ \hline
	\end{tabular}
\end{table}

\par
\begin{figure}[H]
	\includegraphics[width=\linewidth]{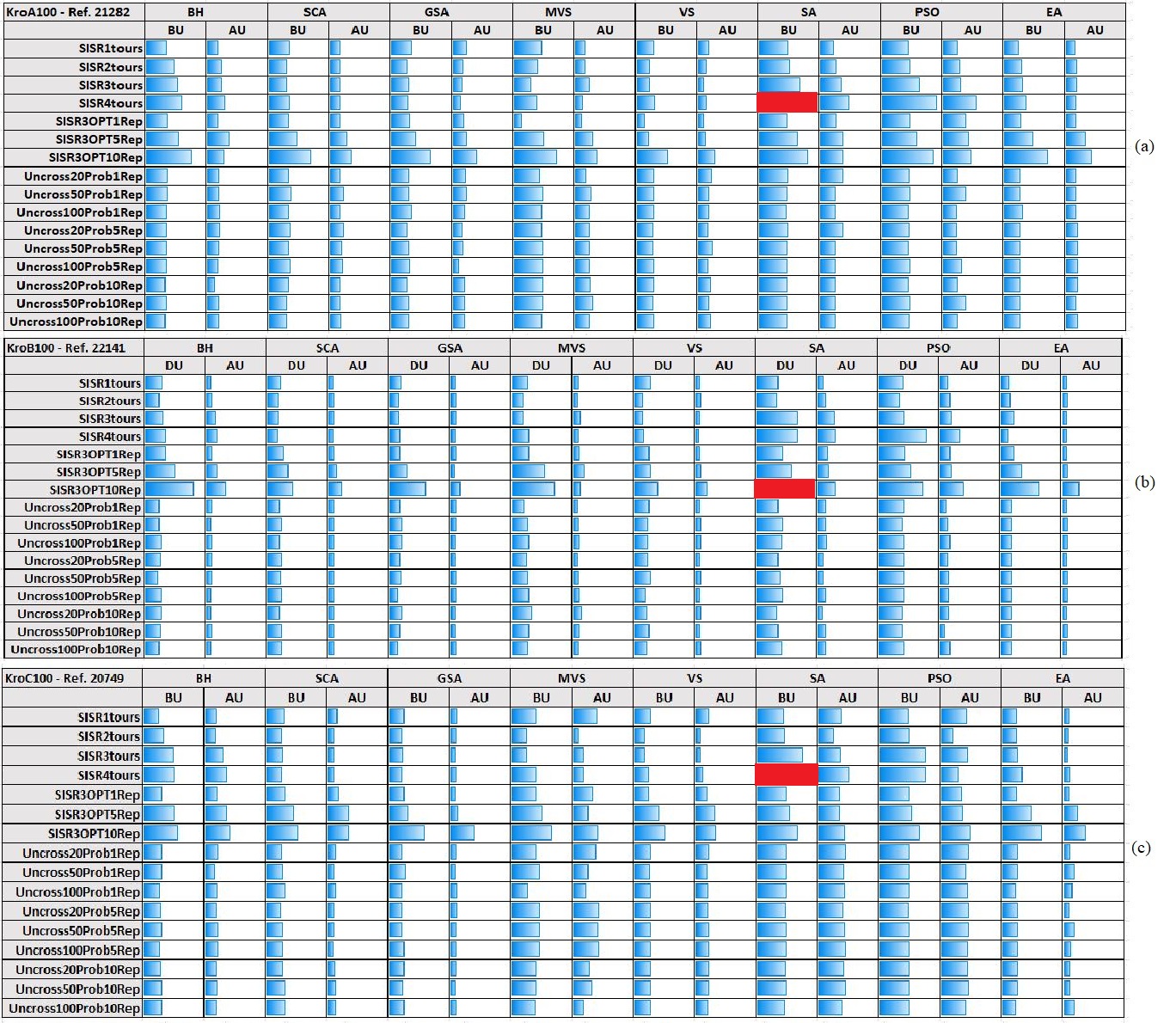}
	\caption{Relative errors obtained with each metaheuristic before (BU) and after (AU) the uncrossing correction heuristic for problems KroA100 (a), KroB100 (b), and KroC100 (c).}
	\label{fig6:dispersaoerros}
\end{figure}

 Figure \ref{fig6:dispersaoerros} shows the amplitude of the normalized errors before and after the final uncross procedure for each problem and all hybrid algorithms evaluated, but TSP-MH and 3OPTYRep. The worst solution obtained before the uncrossing in all tests of each problem is used as the normalization factor: 32.6, 31.24, and 26.03, respectively. The hybridization that provided the worst result is highlighted.The SISRXtours algorithm generated solutions with significant errors for the three problems. However, after applying the final uncrossing heuristic, the solutions with two and three sub-tours are among those with the lowest errors for the three instances. In general, the hybrid algorithms that use the heuristic SISR allied to the uncrossing proved to be more effective in solving the three instances, but not for all metaheuristics tested.

\par Increasing in the number of repetitions of the algorithm 3OPTYRep, in the hybrid SISR3OPTYRep proved ineffective since the error increases as Y increases. However, various reasonable solutions are obtained with Y equal to 1\% of the instance size (number of cities). The fact that the SISR3OPTYRep heuristics modify more and more the solutions provided by metaheuristics as repetitions increase may be the factor that harms the searching process evolution of the metaheuristics, making it more random.

\par The UNCROSSXProbYRep hybridization seems to be almost insensible to the parameters X and Y. For a given instance and metaheuristic, all repetitions result in solutions with almost the same path length and, consequently, the same error. For the instances KroA100 and KroC100, this hybridization provides solutions with a high error. However, some of the best results for KroB100 are obtained with this hybridization. 

\par
\begin{figure}[H]
	\includegraphics[width=\linewidth]{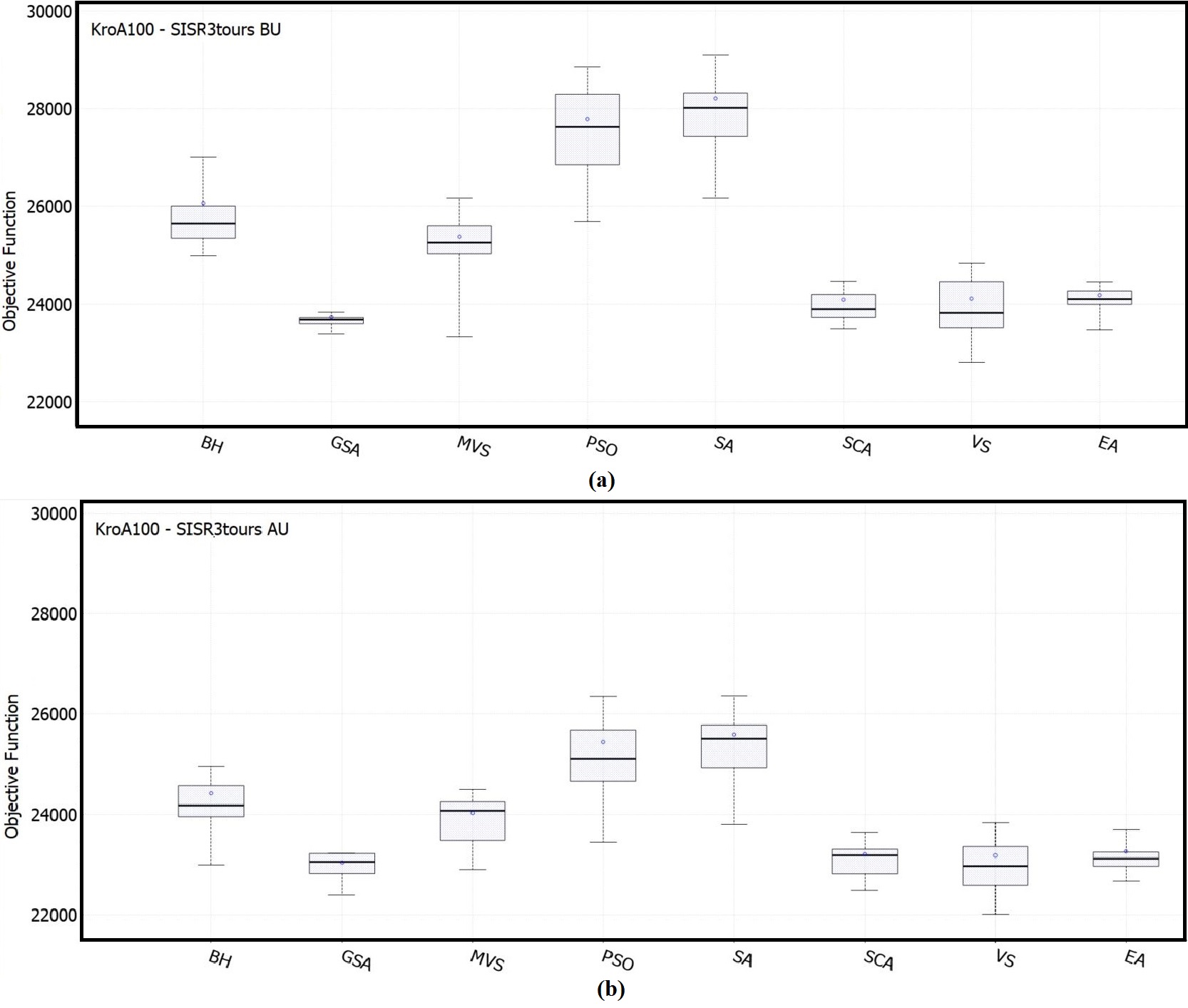}
	\caption{Box plots showing the spread of the best solutions found in the thirty runs executed to solve the instance KroA100 by using the hybrid algorithms metaheuristics/SISR3tours/metaheuristics before correction by the uncrossing heuristic (a) and after the correction by the uncrossing heuristic (b).}
	\label{fig7:boxplot}
\end{figure}

 The algorithm that presented more spread solutions was the PSO. The box plots in Figure \ref{fig7:boxplot} show the spread of the best value found in the thirty repetitions of the hybrid metaheuristics/SISR3tours solving the instance KroA100. In general, the final uncrossing reduces the error associated with the best solution found.

\par
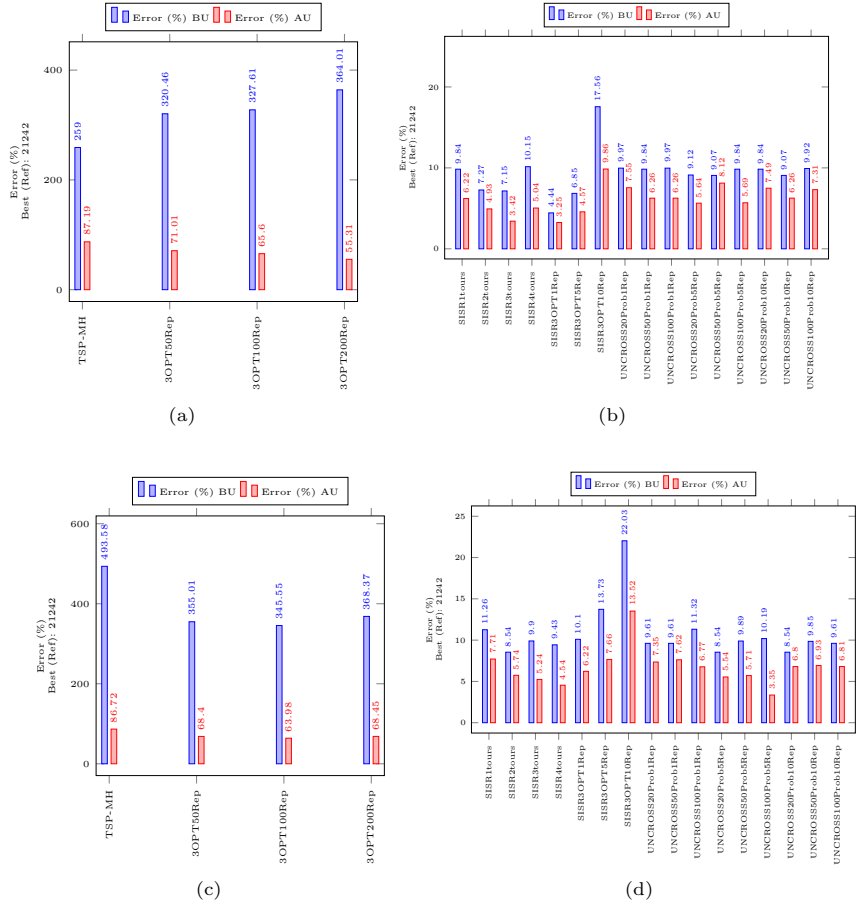
\begin{figure}[H]
	\centering
	\begin{SubFloat}
		{\label{fig:ErroObtidosComAsDiferentesAbordagens1a} }
		\begin{tikzpicture}[scale=0.7]
			\begin{axis}[
				ybar,
				enlarge y limits={abs=3em},
				bar width = {.3em},
				enlargelimits=0.05,
				ylabel={percentage,font=\tiny},
				symbolic x coords={TSP-MH, 3OPT50Rep, 3OPT100Rep, 3OPT200Rep},
				legend style={at={(0.5,1.15)},
					anchor=north,legend columns=-1}, width=7cm, height=6.5cm,
				nodes near coords = {\tiny \pgfmathprintnumber\pgfplotspointmeta},
				nodes near coords align={vertical}, xtick=data,legend style={font=\tiny},
				ylabel style={align=center,font=\tiny}, ylabel=Error (\%)\\Best (Ref): 21242,ymin=0.00, ymax=430.0,every node near coord/.append style={rotate=90, anchor=west,font=\tiny},
				x tick label style={rotate=90,anchor=east,font=\tiny},
				y tick label style={anchor=east,font=\tiny}]
				\legend{Error (\%) BU, Error (\%) AU}
				
				\addplot coordinates  {
					(TSP-MH, 259.0) (3OPT50Rep, 320.46) (3OPT100Rep, 327.61) (3OPT200Rep, 364.01)
				};
				\addplot coordinates {
					(TSP-MH, 87.19) (3OPT50Rep, 71.01) (3OPT100Rep, 65.60) (3OPT200Rep, 55.31)
				};        
			\end{axis}
		\end{tikzpicture}
	\end{SubFloat}
	\vspace{0.2cm}
	\begin{SubFloat}
		{\label{fig:ErroObtidosComAsDiferentesAbordagens1} }
		\begin{tikzpicture}[scale=0.6]
			\begin{axis}[
				ybar,
				enlarge y limits={abs=3em},
				bar width = {.3em},
				enlargelimits=0.05,
				ylabel={percentage,font=\tiny},
				symbolic x coords={SISR1tours, SISR2tours, SISR3tours, SISR4tours, SISR3OPT1Rep, SISR3OPT5Rep, SISR3OPT10Rep, UNCROSS20Prob1Rep, UNCROSS50Prob1Rep,  UNCROSS100Prob1Rep,  UNCROSS20Prob5Rep,  UNCROSS50Prob5Rep,  UNCROSS100Prob5Rep,  UNCROSS20Prob10Rep,  UNCROSS50Prob10Rep,  UNCROSS100Prob10Rep},
				legend style={at={(0.5,1.15)},
					anchor=north,legend columns=-1}, width=10cm, height=6.5cm,
				nodes near coords = {\tiny \pgfmathprintnumber\pgfplotspointmeta},
				nodes near coords align={vertical}, xtick=data,legend style={font=\tiny},
				ylabel style={align=center,font=\tiny}, ylabel=Error (\%)\\Best (Ref): 21242,ymin=0.00, ymax=25.0,every node near coord/.append style={rotate=90, anchor=west,font=\tiny},
				x tick label style={rotate=90,anchor=east,font=\tiny},
				y tick label style={anchor=east,font=\tiny}]
				\legend{Error (\%) BU, Error (\%) AU}
				
				\addplot coordinates  {
					(SISR1tours, 9.84) (SISR2tours, 7.27) (SISR3tours, 7.15) (SISR4tours, 10.15) 
					(SISR3OPT1Rep, 4.44) (SISR3OPT5Rep, 6.85) (SISR3OPT10Rep, 17.56)
					(UNCROSS20Prob1Rep, 9.97) (UNCROSS50Prob1Rep, 9.84) (UNCROSS100Prob1Rep, 9.97)
					(UNCROSS20Prob5Rep, 9.12) (UNCROSS50Prob5Rep, 9.07) (UNCROSS100Prob5Rep, 9.84)
					(UNCROSS20Prob10Rep, 9.84) (UNCROSS50Prob10Rep, 9.07) (UNCROSS100Prob10Rep, 9.92)
				};
				\addplot coordinates {
					(SISR1tours, 6.22) (SISR2tours, 4.93) (SISR3tours, 3.42) (SISR4tours, 5.04)
					(SISR3OPT1Rep, 3.25) (SISR3OPT5Rep, 4.57) (SISR3OPT10Rep, 9.86)
					(UNCROSS20Prob1Rep, 7.55) (UNCROSS50Prob1Rep, 6.26) (UNCROSS100Prob1Rep, 6.26)
					(UNCROSS20Prob5Rep, 5.64) (UNCROSS50Prob5Rep, 8.12) (UNCROSS100Prob5Rep, 5.69)
					(UNCROSS20Prob10Rep, 7.49) (UNCROSS50Prob10Rep, 6.26) (UNCROSS100Prob10Rep, 7.31)
				};        
				
			\end{axis}
		\end{tikzpicture}
	\end{SubFloat}
	\vspace{0.2cm}
	\qquad
	\begin{SubFloat}
		{\label{fig:ErroObtidosComAsDiferentesAbordagens2a} }
		\begin{tikzpicture}[scale=0.7]
			\begin{axis}[
				ybar,
				enlarge y limits={abs=3em},
				bar width = {.3em},
				enlargelimits=0.05,
				ylabel={percentage,font=\tiny},
				symbolic x coords={TSP-MH, 3OPT50Rep, 3OPT100Rep, 3OPT200Rep},
				legend style={at={(0.5,1.15)},
					anchor=north,legend columns=-1},
				nodes near coords = {\tiny \pgfmathprintnumber\pgfplotspointmeta},
				nodes near coords align={vertical}, xtick=data,legend style={font=\tiny}, width=7cm, height=6.5cm,
				ylabel style={align=center,font=\tiny}, ylabel=Error (\%)\\Best (Ref): 21242,ymin=0.00, ymax=590.0,every node near coord/.append style={rotate=90, anchor=west,font=\tiny},
				x tick label style={rotate=90,anchor=east,font=\tiny},
				y tick label style={anchor=east,font=\tiny}]
				\legend{Error (\%) BU, Error (\%) AU}
				
				\addplot coordinates  {
					(TSP-MH, 493.58) (3OPT50Rep, 355.01) (3OPT100Rep, 345.55) (3OPT200Rep, 368.37)
				};
				\addplot coordinates {
					(TSP-MH, 86.72) (3OPT50Rep, 68.40) (3OPT100Rep, 63.98) (3OPT200Rep, 68.45)
				};        
			\end{axis}
		\end{tikzpicture}
	\end{SubFloat}
	\vspace{0.2cm}
	\begin{SubFloat}
		{\label{fig:ErroObtidosComAsDiferentesAbordagens2} }
		\begin{tikzpicture}[scale=0.6]
			\begin{axis}[
				ybar,
				enlarge y limits={abs=3em},
				bar width = {.3em},
				enlargelimits=0.05,
				ylabel={percentage,font=\tiny},
				symbolic x coords={SISR1tours, SISR2tours, SISR3tours, SISR4tours, SISR3OPT1Rep, SISR3OPT5Rep, SISR3OPT10Rep, UNCROSS20Prob1Rep, UNCROSS50Prob1Rep,  UNCROSS100Prob1Rep,  UNCROSS20Prob5Rep,  UNCROSS50Prob5Rep,  UNCROSS100Prob5Rep,  UNCROSS20Prob10Rep,  UNCROSS50Prob10Rep,  UNCROSS100Prob10Rep},
				legend style={at={(0.5,1.15)},
					anchor=north,legend columns=-1},
				nodes near coords = {\tiny \pgfmathprintnumber\pgfplotspointmeta},
				nodes near coords align={vertical}, xtick=data,legend style={font=\tiny}, width=10cm, height=6.6cm,
				ylabel style={align=center,font=\tiny}, ylabel=Error (\%)\\Best (Ref): 21242,ymin=0.00, ymax=25.0,every node near coord/.append style={rotate=90, anchor=west,font=\tiny},
				x tick label style={rotate=90,anchor=east,font=\tiny},
				y tick label style={anchor=east,font=\tiny}]
				\legend{Error (\%) BU, Error (\%) AU}
				
				\addplot coordinates  {
					(SISR1tours, 11.26) (SISR2tours, 8.54) (SISR3tours, 9.90) (SISR4tours, 9.43)
					(SISR3OPT1Rep, 10.10) (SISR3OPT5Rep, 13.73) (SISR3OPT10Rep, 22.03)
					(UNCROSS20Prob1Rep, 9.61) (UNCROSS50Prob1Rep, 9.61) (UNCROSS100Prob1Rep, 11.32)
					(UNCROSS20Prob5Rep, 8.54) (UNCROSS50Prob5Rep, 9.89) (UNCROSS100Prob5Rep, 10.19)
					(UNCROSS20Prob10Rep, 8.54) (UNCROSS50Prob10Rep, 9.85) (UNCROSS100Prob10Rep, 9.61)
				};
				\addplot coordinates {
					(SISR1tours, 7.71) (SISR2tours, 5.74) (SISR3tours, 5.24) (SISR4tours, 4.54)
					(SISR3OPT1Rep, 6.22) (SISR3OPT5Rep, 7.66) (SISR3OPT10Rep, 13.52)
					(UNCROSS20Prob1Rep, 7.35) (UNCROSS50Prob1Rep, 7.62) (UNCROSS100Prob1Rep, 6.77)
					(UNCROSS20Prob5Rep, 5.54) (UNCROSS50Prob5Rep, 5.71) (UNCROSS100Prob5Rep, 3.35)
					(UNCROSS20Prob10Rep, 6.80) (UNCROSS50Prob10Rep, 6.93) (UNCROSS100Prob10Rep, 6.81)
				};        
			\end{axis}
		\end{tikzpicture}
	\end{SubFloat}
	\caption{Errors of the solutions obtained with the hybrid algorithms using VS (a) and GSA (b), for KroA100 instance, before (BU) and after correction with the uncrossing heuristic (AU).}
	\label{fig8:ErroObtidos}
\end{figure}

 In most cases where the SISR heuristic was used, such as the 3OPT50Rep, the solutions obtained present errors smaller than 8\% after the final uncrossing verification, a value more than ten times smaller than those obtained without using the SISR heuristic. Figure \ref{fig8:ErroObtidos} shows the errors obtained by the hybrid algorithms with two particular metaheuristics: VS (Figure \ref{fig8:ErroObtidos}.a and Figure \ref{fig8:ErroObtidos}.b) and GSA (Figure \ref{fig8:ErroObtidos}.c and Figure \ref{fig8:ErroObtidos}.d), solving the KroA100 instance before and after the uncrossing corrections. These hybrid algorithms found the best solutions in this study regarding the optimal solution recorded in the literature for the KroA100 problem. As expressed previously, TSP-MH, 3OPT50Rep, 3OPT100Rep and 3OPT200Rep hybrid algorithms present the worst results, Figures \ref{fig8:ErroObtidos}.a and \ref{fig8:ErroObtidos}.c. Even with the final uncrossing, the errors are superior to 55\%. For the other hybridizations, Figures \ref{fig8:ErroObtidos}.b and \ref{fig8:ErroObtidos}.d, which include the SISR, the mean error reduction is about 34\% after the uncrossing. In some cases, the reduction achieves more than 60\%. The error of the best solutions obtained are 3.25\% using VS/SISR3OPT1REP hybrid optimization and 3.35\% using GSA/UNCROSS100Prob5Rep. Notice that the hybrid algorithms VS/SISR3OPT10Rep and GSA/SISR3OPT10Rep show errors superior to 8\%.

\par Generally, the VS, MVS, and BH metaheuristics get the best results hybridized either with the SISR heuristic with sub-tours or with the SISR3OPT heuristic. These three metaheuristics have searching procedures for the optimum that attract the solutions to regions closer and closer to a centre (the best solution so far).

\par SCA, PSO and SA metaheuristics obtained the best solutions with the UNCROSSXProbYRep algorithm, which also uses the SISR. The solutions obtained with the SCA have smaller errors than those of the other two algorithms. In general, the solutions obtained by the hybrid algorithms with PSO and SA in this experiment set have errors higher than the other hybridizations. Still, the hybridizations SA/Uncross20Prob10Rep and PSO/Uncross50Prob10Rep have found solutions for KroB100 with errors close to the best solutions.

\par All hybridizations using the GSA metaheuristic have presented good solutions in all tests and a weak dependency on the hybridization regarding KroB100 and KroC100, Figure \ref{fig6:dispersaoerros}. 

\par The evolutionary algorithm obtained the best solution for KroC100 when hybridized with the SISR3tours. Moreover, the solutions present errors close to the obtained for the best solution found for the three instances, and the errors also show a weak dependency on hybridization.

\par Figure \ref{fig9:TempoMedioGastoPelasMH-Todos} shows the average number of objective function calls (OFC) spent by each metaheuristic to find the best solution in all three instances, expressed in terms of a percentage of the total number of objective function calls. Systematically, hybridizations with the SCA and SA metaheuristics achieve the final solution earlier than the others, but frequently the solutions get stuck in local minima. The best solution is generally found using the VS metaheuristic; however, it takes more function calls to obtain the solution, which could suggest that better results would be achieved with more function calls. However, tests show that the results do not change even with up to 1E6 objective function calls. MVS present a similar convergence behaviour, giving good solutions but spending more objective function calls.
\begin{figure}[H]
	\centering
	\begin{SubFloat}
		{\label{fig:TempoMedioGastoPelasMHKro1} }
		\begin{tikzpicture}[scale=0.73]
			\begin{axis}[legend style={at={(1.01,0.5)},anchor=west,font=\tiny},
				symbolic x coords={TSP-MH, 3OPT50Rep, 3OPT100Rep, 3OPT200Rep, SISR1tours, SISR2tours, SISR3tours, SISR4tours, SISR3OPT1Rep, SISR3OPT5Rep, SISR3OPT10Rep, UNCROSS20Prob1Rep, UNCROSS50Prob1Rep,  UNCROSS100Prob1Rep,  UNCROSS20Prob5Rep,  UNCROSS50Prob5Rep,  UNCROSS100Prob5Rep,  UNCROSS20Prob10Rep,  UNCROSS50Prob10Rep,  UNCROSS100Prob10Rep}, xtick=data,
				ylabel={OFC (\%)},ymin=0.10, ymax=100,
				x tick label style={rotate=90,anchor=east,font=\tiny},width=15cm,height=6.5cm,font=\tiny
				]
				\addlegendentry{BH}
				\addplot[mark=*,thick,blue] coordinates {
					(TSP-MH, 91.63) (3OPT50Rep, 45.26) (3OPT100Rep, 54.98) (3OPT200Rep, 42.61)
					(SISR1tours, 60.42) (SISR2tours, 53.56) (SISR3tours, 55.66) (SISR4tours, 53.22)
					(SISR3OPT1Rep, 60.86) (SISR3OPT5Rep, 62.09) (SISR3OPT10Rep, 57.70)
					(UNCROSS20Prob1Rep, 57.38) (UNCROSS50Prob1Rep, 63.31) (UNCROSS100Prob1Rep, 58.03)
					(UNCROSS20Prob5Rep, 65.61) (UNCROSS50Prob5Rep, 60.05) (UNCROSS100Prob5Rep, 64.89)
					(UNCROSS20Prob10Rep, 62.57) (UNCROSS50Prob10Rep, 61.08) (UNCROSS100Prob10Rep, 59.07)
				};
				
				\addlegendentry{GSA}
				\addplot[mark=diamond*,thick,red] coordinates {
					(TSP-MH, 47.43) (3OPT50Rep, 66.21) (3OPT100Rep, 52.50) (3OPT200Rep, 62.25)
					(SISR1tours, 43.00) (SISR2tours, 55.21) (SISR3tours, 58.77) (SISR4tours, 56.45) 
					(SISR3OPT1Rep, 49.36) (SISR3OPT5Rep, 53.32) (SISR3OPT10Rep, 56.27)
					(UNCROSS20Prob1Rep, 59.32) (UNCROSS50Prob1Rep, 60.60) (UNCROSS100Prob1Rep, 50.26)
					(UNCROSS20Prob5Rep, 41.56) (UNCROSS50Prob5Rep, 60.81) (UNCROSS100Prob5Rep, 50.85)
					(UNCROSS20Prob10Rep, 45.77) (UNCROSS50Prob10Rep, 45.13) (UNCROSS100Prob10Rep, 48.62)
				};
				
				\addlegendentry{MVS}
				\addplot[mark=square,thick,green] coordinates {
					(TSP-MH, 78.6) (3OPT50Rep, 79.6) (3OPT100Rep, 70.5) (3OPT200Rep, 53.8)
					(SISR1tours, 62.8) (SISR2tours, 84.9) (SISR3tours, 85.7) (SISR4tours, 85.3) 
					(SISR3OPT1Rep, 90.9) (SISR3OPT5Rep, 84.0) (SISR3OPT10Rep, 80.3)
					(UNCROSS20Prob1Rep, 64.3) (UNCROSS50Prob1Rep, 71.7) (UNCROSS100Prob1Rep, 70.2)
					(UNCROSS20Prob5Rep, 57.9) (UNCROSS50Prob5Rep, 70.4) (UNCROSS100Prob5Rep, 75.1)
					(UNCROSS20Prob10Rep, 69.5) (UNCROSS50Prob10Rep, 61.7) (UNCROSS100Prob10Rep, 61.2)
				};
				
				\addlegendentry{PSO}
				\addplot[mark=halfcircle,thick,orange] coordinates {
					(TSP-MH, 63.35) (3OPT50Rep, 45.21) (3OPT100Rep, 48.41) (3OPT200Rep, 49.80)
					(SISR1tours, 48.15) (SISR2tours, 46.11) (SISR3tours, 48.98) (SISR4tours, 59.35)
					(SISR3OPT1Rep, 48.07) (SISR3OPT5Rep, 37.06) (SISR3OPT10Rep, 50.09)
					(UNCROSS20Prob1Rep, 35.02) (UNCROSS50Prob1Rep, 33.24) (UNCROSS100Prob1Rep, 32.95)
					(UNCROSS20Prob5Rep, 41.44) (UNCROSS50Prob5Rep, 33.37) (UNCROSS100Prob5Rep, 31.66)
					(UNCROSS20Prob10Rep, 40.51) (UNCROSS50Prob10Rep, 35.96) (UNCROSS100Prob10Rep, 38.87)
				};
				
				\addlegendentry{SA}
				\addplot[mark=halfcircle*,thick,gray] coordinates {
					(TSP-MH, 48.95) (3OPT50Rep, 55.96) (3OPT100Rep, 39.60) (3OPT200Rep, 55.96)
					(SISR1tours, 28.85) (SISR2tours, 56.44) (SISR3tours, 50.82) (SISR4tours, 44.67)
					(SISR3OPT1Rep, 48.56) (SISR3OPT5Rep, 48.69) (SISR3OPT10Rep, 48.10)
					(UNCROSS20Prob1Rep, 37.71) (UNCROSS50Prob1Rep, 21.76) (UNCROSS100Prob1Rep, 27.65)
					(UNCROSS20Prob5Rep, 31.15) (UNCROSS50Prob5Rep, 30.87) (UNCROSS100Prob5Rep, 34.98)
					(UNCROSS20Prob10Rep, 28.98) (UNCROSS50Prob10Rep, 27.23) (UNCROSS100Prob10Rep, 29.09)
				};
				
				\addlegendentry{SCA}
				\addplot[mark=square*,thick,pink] coordinates {
					(TSP-MH, 47.50) (3OPT50Rep, 56.16) (3OPT100Rep, 60.08) (3OPT200Rep, 55.80)
					(SISR1tours, 35.69) (SISR2tours, 35.31) (SISR3tours, 29.71) (SISR4tours, 32.05)
					(SISR3OPT1Rep, 26.83) (SISR3OPT5Rep, 29.22) (SISR3OPT10Rep, 37.30)
					(UNCROSS20Prob1Rep, 36.44) (UNCROSS50Prob1Rep, 32.57) (UNCROSS100Prob1Rep, 33.15)
					(UNCROSS20Prob5Rep, 33.06) (UNCROSS50Prob5Rep, 34.61) (UNCROSS100Prob5Rep, 17.54)
					(UNCROSS20Prob10Rep, 34.81) (UNCROSS50Prob10Rep, 27.88) (UNCROSS100Prob10Rep, 25.18)
				};
				
				\addlegendentry{VS}
				\addplot[mark=diamond,thick,brown] coordinates {
					(TSP-MH, 68.80) (3OPT50Rep, 86.84) (3OPT100Rep, 75.75) (3OPT200Rep, 66.30)
					(SISR1tours, 45.08) (SISR2tours, 71.13) (SISR3tours, 73.49) (SISR4tours, 73.82) 
					(SISR3OPT1Rep, 71.64) (SISR3OPT5Rep, 93.07) (SISR3OPT10Rep, 83.61)
					(UNCROSS20Prob1Rep, 54.05) (UNCROSS50Prob1Rep, 51.53) (UNCROSS100Prob1Rep, 47.26)
					(UNCROSS20Prob5Rep, 49.00) (UNCROSS50Prob5Rep, 52.93) (UNCROSS100Prob5Rep, 48.18)
					(UNCROSS20Prob10Rep, 46.18) (UNCROSS50Prob10Rep, 44.75) (UNCROSS100Prob10Rep, 50.69)
				};
				
				\addlegendentry{EA}
				\addplot[mark=diamond,thick,black] coordinates {
					(TSP-MH, 0.0) (3OPT50Rep, 46.51) (3OPT100Rep, 56.60) (3OPT200Rep, 43.91)
					(SISR1tours, 41.42) (SISR2tours, 54.34) (SISR3tours, 54.31) (SISR4tours, 47.73) 
					(SISR3OPT1Rep, 42.04) (SISR3OPT5Rep, 46.88) (SISR3OPT10Rep, 52.62)
					(UNCROSS20Prob1Rep, 37.74) (UNCROSS50Prob1Rep, 36.02) (UNCROSS100Prob1Rep, 35.44)
					(UNCROSS20Prob5Rep, 40.98) (UNCROSS50Prob5Rep, 41.65) (UNCROSS100Prob5Rep, 29.83)
					(UNCROSS20Prob10Rep, 44.42) (UNCROSS50Prob10Rep, 38.19) (UNCROSS100Prob10Rep, 34.78)
				};
			\end{axis}
		\end{tikzpicture}
	\end{SubFloat}
	\begin{SubFloat}
		{\label{fig:TempoMedioGastoPelasMHKro2} }	
		\begin{tikzpicture}[scale=0.73]
			\begin{axis}[legend style={at={(1.01,0.5)},anchor=west,font=\tiny},
				symbolic x coords={TSP-MH, 3OPT50Rep, 3OPT100Rep, 3OPT200Rep, SISR1tours, SISR2tours, SISR3tours, SISR4tours, SISR3OPT1Rep, SISR3OPT5Rep, 
					SISR3OPT10Rep, UNCROSS20Prob1Rep, UNCROSS50Prob1Rep,  UNCROSS100Prob1Rep,  UNCROSS20Prob5Rep,  UNCROSS50Prob5Rep,  UNCROSS100Prob5Rep,  UNCROSS20Prob10Rep,  UNCROSS50Prob10Rep,  UNCROSS100Prob10Rep}, xtick=data,
				ylabel={OFC (\%)},ymin=0.10, ymax=100,
				x tick label style={rotate=90,anchor=east,font=\tiny},width=15cm,height=6.5cm,font=\tiny
				]
				\addlegendentry{BH}
				\addplot[mark=*,thick,blue] coordinates {
					(TSP-MH, 0.9) (3OPT50Rep, 46.45) (3OPT100Rep, 51.31) (3OPT200Rep, 46.73)
					(SISR1tours, 58.67) (SISR2tours, 50.22) (SISR3tours, 46.13) (SISR4tours, 57.69)
					(SISR3OPT1Rep, 57.11) (SISR3OPT5Rep, 65.76) (SISR3OPT10Rep, 56.49)
					(UNCROSS20Prob1Rep, 50.55) (UNCROSS50Prob1Rep, 61.31) (UNCROSS100Prob1Rep, 52.35)
					(UNCROSS20Prob5Rep, 53.87) (UNCROSS50Prob5Rep, 55.89) (UNCROSS100Prob5Rep, 52.07)
					(UNCROSS20Prob10Rep, 60.58) (UNCROSS50Prob10Rep, 51.22) (UNCROSS100Prob10Rep, 51.93)
				};
				
				\addlegendentry{GSA}
				\addplot[mark=diamond*,thick,red] coordinates {
					(TSP-MH, 0.15) (3OPT50Rep, 60.99) (3OPT100Rep, 50.48) (3OPT200Rep, 57.63)
					(SISR1tours, 44.24) (SISR2tours, 59.40) (SISR3tours, 60.55) (SISR4tours, 53.61) 
					(SISR3OPT1Rep, 54.10) (SISR3OPT5Rep, 56.60) (SISR3OPT10Rep, 54.42)
					(UNCROSS20Prob1Rep, 54.94) (UNCROSS50Prob1Rep, 56.55) (UNCROSS100Prob1Rep, 44.58)
					(UNCROSS20Prob5Rep, 51.87) (UNCROSS50Prob5Rep, 38.58) (UNCROSS100Prob5Rep, 49.64)
					(UNCROSS20Prob10Rep, 44.11) (UNCROSS50Prob10Rep, 46.44) (UNCROSS100Prob10Rep, 43.25)
				};
				
				\addlegendentry{MVS}
				\addplot[mark=square,thick,green] coordinates {
					(TSP-MH, 78.5) (3OPT50Rep, 77.3) (3OPT100Rep, 67.9) (3OPT200Rep, 46.2)
					(SISR1tours, 69.8) (SISR2tours, 81.5) (SISR3tours, 79.5) (SISR4tours, 81.3) 
					(SISR3OPT1Rep, 85.7) (SISR3OPT5Rep, 85.3) (SISR3OPT10Rep, 77.4)
					(UNCROSS20Prob1Rep, 58.2) (UNCROSS50Prob1Rep, 68.8) (UNCROSS100Prob1Rep, 70.2)
					(UNCROSS20Prob5Rep, 72.9) (UNCROSS50Prob5Rep, 58.4) (UNCROSS100Prob5Rep, 62.7)
					(UNCROSS20Prob10Rep, 73.9) (UNCROSS50Prob10Rep, 68.3) (UNCROSS100Prob10Rep, 68.0)
				};
				
				\addlegendentry{PSO}
				\addplot[mark=halfcircle,thick,orange] coordinates {
					(TSP-MH, 0.56) (3OPT50Rep, 44.85) (3OPT100Rep, 58.22) (3OPT200Rep, 47.57)
					(SISR1tours, 49.03) (SISR2tours, 59.16) (SISR3tours, 45.04) (SISR4tours, 52.57)
					(SISR3OPT1Rep, 50.86) (SISR3OPT5Rep, 49.32) (SISR3OPT10Rep, 36.29)
					(UNCROSS20Prob1Rep, 52.05) (UNCROSS50Prob1Rep, 61.87) (UNCROSS100Prob1Rep, 44.17)
					(UNCROSS20Prob5Rep, 48.00) (UNCROSS50Prob5Rep, 42.08) (UNCROSS100Prob5Rep, 37.74)
					(UNCROSS20Prob10Rep, 46.28) (UNCROSS50Prob10Rep, 38.44) (UNCROSS100Prob10Rep, 39.94)
				};
				
				\addlegendentry{SA}
				\addplot[mark=halfcircle*,thick,gray] coordinates {
					(TSP-MH, 0.52) (3OPT50Rep, 57.24) (3OPT100Rep, 52.98) (3OPT200Rep, 44.30)
					(SISR1tours, 31.82) (SISR2tours, 60.48) (SISR3tours, 44.61) (SISR4tours, 40.11)
					(SISR3OPT1Rep, 40.59) (SISR3OPT5Rep, 49.19) (SISR3OPT10Rep, 41.81)
					(UNCROSS20Prob1Rep, 43.20) (UNCROSS50Prob1Rep, 48.60) (UNCROSS100Prob1Rep, 48.53)
					(UNCROSS20Prob5Rep, 45.75) (UNCROSS50Prob5Rep, 44.78) (UNCROSS100Prob5Rep, 39.67)
					(UNCROSS20Prob10Rep, 50.64) (UNCROSS50Prob10Rep, 42.81) (UNCROSS100Prob10Rep, 59.47)
				};
				
				\addlegendentry{SCA}
				\addplot[mark=square*,thick,pink] coordinates {
					(TSP-MH, 0.44) (3OPT50Rep, 49.13) (3OPT100Rep, 51.82) (3OPT200Rep, 50.59)
					(SISR1tours, 20.29) (SISR2tours, 25.32) (SISR3tours, 22.76) (SISR4tours, 20.28)
					(SISR3OPT1Rep, 34.12) (SISR3OPT5Rep, 29.65) (SISR3OPT10Rep, 32.16)
					(UNCROSS20Prob1Rep, 24.80) (UNCROSS50Prob1Rep, 31.30) (UNCROSS100Prob1Rep, 33.15)
					(UNCROSS20Prob5Rep, 29.37) (UNCROSS50Prob5Rep, 29.87) (UNCROSS100Prob5Rep, 24.96)
					(UNCROSS20Prob10Rep, 37.00) (UNCROSS50Prob10Rep, 35.04) (UNCROSS100Prob10Rep, 27.64)
				};
				
				\addlegendentry{VS}
				\addplot[mark=diamond,thick,brown] coordinates {
					(TSP-MH, 0.79) (3OPT50Rep, 85.83) (3OPT100Rep, 80.56) (3OPT200Rep, 74.15)
					(SISR1tours, 70.76) (SISR2tours, 72.58) (SISR3tours, 75.12) (SISR4tours, 74.54) 
					(SISR3OPT1Rep, 76.94) (SISR3OPT5Rep, 87.16) (SISR3OPT10Rep, 86.87)
					(UNCROSS20Prob1Rep, 69.78) (UNCROSS50Prob1Rep, 71.71) (UNCROSS100Prob1Rep, 70.50)
					(UNCROSS20Prob5Rep, 71.11) (UNCROSS50Prob5Rep, 73.99) (UNCROSS100Prob5Rep, 75.64)
					(UNCROSS20Prob10Rep, 72.68) (UNCROSS50Prob10Rep, 68.02) (UNCROSS100Prob10Rep, 74.09)
				};
				
				\addlegendentry{EA}
				\addplot[mark=diamond,thick,black] coordinates {
					(TSP-MH, 0.0) (3OPT50Rep, 49.17) (3OPT100Rep, 44.70) (3OPT200Rep, 55.10)
					(SISR1tours, 54.72) (SISR2tours, 52.22) (SISR3tours, 48.51) (SISR4tours, 59.49) 
					(SISR3OPT1Rep, 46.84) (SISR3OPT5Rep, 47.81) (SISR3OPT10Rep, 50.65)
					(UNCROSS20Prob1Rep, 44.41) (UNCROSS50Prob1Rep, 52.49) (UNCROSS100Prob1Rep, 48.17)
					(UNCROSS20Prob5Rep, 59.77) (UNCROSS50Prob5Rep, 45.53) (UNCROSS100Prob5Rep, 46.93)
					(UNCROSS20Prob10Rep, 40.81) (UNCROSS50Prob10Rep, 45.07) (UNCROSS100Prob10Rep, 55.85)
				};
				
			\end{axis}
		\end{tikzpicture}
	\end{SubFloat}
	\begin{SubFloat}
		{\label{fig:TempoMedioGastoPelasMHKro3} }
		\begin{tikzpicture}[scale=0.73]
			\begin{axis}[legend style={at={(1.01,0.5)},anchor=west,font=\tiny},
				symbolic x coords={TSP-MH, 3OPT50Rep, 3OPT100Rep, 3OPT200Rep, SISR1tours, SISR2tours, SISR3tours, SISR4tours, SISR3OPT1Rep, SISR3OPT5Rep, 
					SISR3OPT10Rep, UNCROSS20Prob1Rep, UNCROSS50Prob1Rep,  UNCROSS100Prob1Rep,  UNCROSS20Prob5Rep,  UNCROSS50Prob5Rep,  UNCROSS100Prob5Rep,  
					UNCROSS20Prob10Rep,  UNCROSS50Prob10Rep,  UNCROSS100Prob10Rep}, xtick=data,
				ylabel={OFC (\%)},ymin=0.10, ymax=100,
				x tick label style={rotate=90,anchor=east,font=\tiny},width=15cm,height=6.5cm,font=\tiny
				]
				\addlegendentry{BH}
				\addplot[mark=*,thick,blue] coordinates {
					(TSP-MH, 93.63) (3OPT50Rep, 51.09) (3OPT100Rep, 47.06) (3OPT200Rep, 46.71)
					(SISR1tours, 43.49) (SISR2tours, 55.34) (SISR3tours, 42.33) (SISR4tours, 51.83)
					(SISR3OPT1Rep, 51.18) (SISR3OPT5Rep, 55.30) (SISR3OPT10Rep, 54.37)
					(UNCROSS20Prob1Rep, 57.32) (UNCROSS50Prob1Rep, 64.26) (UNCROSS100Prob1Rep, 55.88)
					(UNCROSS20Prob5Rep, 52.92) (UNCROSS50Prob5Rep, 62.25) (UNCROSS100Prob5Rep, 58.25)
					(UNCROSS20Prob10Rep, 58.46) (UNCROSS50Prob10Rep, 64.25) (UNCROSS100Prob10Rep, 57.17)
				};

				\addlegendentry{GSA}
				\addplot[mark=diamond*,thick,red] coordinates {
					(TSP-MH, 39.35) (3OPT50Rep, 48.37) (3OPT100Rep, 68.31) (3OPT200Rep, 59.48)
					(SISR1tours, 47.76) (SISR2tours, 51.31) (SISR3tours, 48.54) (SISR4tours, 54.24) 
					(SISR3OPT1Rep, 56.21) (SISR3OPT5Rep, 55.89) (SISR3OPT10Rep, 45.72)
					(UNCROSS20Prob1Rep, 45.76) (UNCROSS50Prob1Rep, 51.85) (UNCROSS100Prob1Rep, 44.52)
					(UNCROSS20Prob5Rep, 44.93) (UNCROSS50Prob5Rep, 44.22) (UNCROSS100Prob5Rep, 57.69)
					(UNCROSS20Prob10Rep, 56.60) (UNCROSS50Prob10Rep, 49.15) (UNCROSS100Prob10Rep, 44.51)
				};
				
				\addlegendentry{MVS}
				\addplot[mark=square,thick,green] coordinates {
					(TSP-MH, 76.2) (3OPT50Rep, 72.7) (3OPT100Rep, 63.1) (3OPT200Rep, 51.4)
					(SISR1tours, 56.0) (SISR2tours, 78.5) (SISR3tours, 80.8) (SISR4tours, 81.9) 
					(SISR3OPT1Rep, 78.8) (SISR3OPT5Rep, 85.1) (SISR3OPT10Rep, 72.4)
					(UNCROSS20Prob1Rep, 53.5) (UNCROSS50Prob1Rep, 61.6) (UNCROSS100Prob1Rep, 55.2)
					(UNCROSS20Prob5Rep, 46.0) (UNCROSS50Prob5Rep, 55.3) (UNCROSS100Prob5Rep, 60.9)
					(UNCROSS20Prob10Rep, 65.5) (UNCROSS50Prob10Rep, 63.3) (UNCROSS100Prob10Rep, 59.7)
				};
				
				\addlegendentry{PSO}
				\addplot[mark=halfcircle,thick,orange] coordinates {
					(TSP-MH, 51.37) (3OPT50Rep, 37.54) (3OPT100Rep, 55.18) (3OPT200Rep, 49.16)
					(SISR1tours, 25.93) (SISR2tours, 42.31) (SISR3tours, 45.03) (SISR4tours, 39.40)
					(SISR3OPT1Rep, 37.88) (SISR3OPT5Rep, 55.49) (SISR3OPT10Rep, 50.44)
					(UNCROSS20Prob1Rep, 30.59) (UNCROSS50Prob1Rep, 28.62) (UNCROSS100Prob1Rep, 31.22)
					(UNCROSS20Prob5Rep, 43.47) (UNCROSS50Prob5Rep, 37.82) (UNCROSS100Prob5Rep, 36.69)
					(UNCROSS20Prob10Rep, 37.23) (UNCROSS50Prob10Rep, 30.07) (UNCROSS100Prob10Rep, 29.33)
				};
				
				\addlegendentry{SA}
				\addplot[mark=halfcircle*,thick,gray] coordinates {
					(TSP-MH, 50.82) (3OPT50Rep, 53.53) (3OPT100Rep, 48.41) (3OPT200Rep, 42.13)
					(SISR1tours, 15.98) (SISR2tours, 42.19) (SISR3tours, 53.85) (SISR4tours, 55.12)
					(SISR3OPT1Rep, 36.28) (SISR3OPT5Rep, 48.17) (SISR3OPT10Rep, 44.34)
					(UNCROSS20Prob1Rep, 26.99) (UNCROSS50Prob1Rep, 33.30) (UNCROSS100Prob1Rep, 30.78)
					(UNCROSS20Prob5Rep, 30.98) (UNCROSS50Prob5Rep, 36.09) (UNCROSS100Prob5Rep, 34.71)
					(UNCROSS20Prob10Rep, 31.65) (UNCROSS50Prob10Rep, 38.62) (UNCROSS100Prob10Rep, 26.61)
				};
				
				\addlegendentry{SCA}
				\addplot[mark=square*,thick,pink] coordinates {
					(TSP-MH, 48.95) (3OPT50Rep, 59.70) (3OPT100Rep, 56.02) (3OPT200Rep, 52.41)
					(SISR1tours, 34.09) (SISR2tours, 31.67) (SISR3tours, 29.25) (SISR4tours, 31.03)
					(SISR3OPT1Rep, 29.99) (SISR3OPT5Rep, 31.50) (SISR3OPT10Rep, 33.75)
					(UNCROSS20Prob1Rep, 29.41) (UNCROSS50Prob1Rep, 30.99) (UNCROSS100Prob1Rep, 28.17)
					(UNCROSS20Prob5Rep, 38.03) (UNCROSS50Prob5Rep, 30.23) (UNCROSS100Prob5Rep, 44.03)
					(UNCROSS20Prob10Rep, 39.22) (UNCROSS50Prob10Rep, 32.84) (UNCROSS100Prob10Rep, 35.17)
				};
				
				\addlegendentry{VS}
				\addplot[mark=diamond,thick,brown] coordinates {
					(TSP-MH, 64.43) (3OPT50Rep, 83.54) (3OPT100Rep, 82.29) (3OPT200Rep, 69.78)
					(SISR1tours, 67.03) (SISR2tours, 72.89) (SISR3tours, 75.71) (SISR4tours, 75.29) 
					(SISR3OPT1Rep, 77.00) (SISR3OPT5Rep, 79.58) (SISR3OPT10Rep, 82.17)
					(UNCROSS20Prob1Rep, 67.95) (UNCROSS50Prob1Rep, 66.35) (UNCROSS100Prob1Rep, 66.82)
					(UNCROSS20Prob5Rep, 67.52) (UNCROSS50Prob5Rep, 66.97) (UNCROSS100Prob5Rep, 67.44)
					(UNCROSS20Prob10Rep, 66.94) (UNCROSS50Prob10Rep, 68.17) (UNCROSS100Prob10Rep, 67.10)
				};
				
				\addlegendentry{EA}
				\addplot[mark=diamond,thick,black] coordinates {
					(TSP-MH, 0.0) (3OPT50Rep, 47.13) (3OPT100Rep, 51.32) (3OPT200Rep, 56.17)
					(SISR1tours, 43.79) (SISR2tours, 51.29) (SISR3tours, 47.43) (SISR4tours, 54.93) 
					(SISR3OPT1Rep, 49.73) (SISR3OPT5Rep, 52.99) (SISR3OPT10Rep, 49.48)
					(UNCROSS20Prob1Rep, 43.88) (UNCROSS50Prob1Rep, 41.89) (UNCROSS100Prob1Rep, 48.43)
					(UNCROSS20Prob5Rep, 39.72) (UNCROSS50Prob5Rep, 39.25) (UNCROSS100Prob5Rep, 45.49)
					(UNCROSS20Prob10Rep, 54.89) (UNCROSS50Prob10Rep, 47.02) (UNCROSS100Prob10Rep, 42.56)
				};
			\end{axis}
		\end{tikzpicture}
	\end{SubFloat}
	\caption{Average objective function calls (OFC) (\%) required to find the best solution in problems KroA100 (a), KroB100 (b), and KroC100 (c).}
	\label{fig9:TempoMedioGastoPelasMH-Todos}
\end{figure}

\par In the KroB100 problem, all metaheuristics with the correction strategy and no hybridization, named generically as TSP-MH, converged very early, Figure \ref{fig9:TempoMedioGastoPelasMH-Todos}. All hybrid algorithms, except the ones using MVS, converge with less than 1\% of objective function calls. However, the solutions get stuck in local minima, presenting errors greater than 57.9\% compared to the known optimal solution. Only the EA metaheuristic had the same behaviour in solving the other two instances.

\par The results present evidence of heuristics that effectively found good solutions to KroX100 family problems. No hybridization technique showed a clear advantage in solving the different instances, but the hybridization of metaheuristics with the 3OPT algorithm and without the SISR could be discarded. The best hybrid algorithms identified in this experiment were SISR3OPTYRep, with Y=1\%, SISRXtours, UncrossXProbYRep, with X = 20 and 50\% and Y=1 and 5\%. The BH, EA, GSA, MVS, SCA, and VS hybridizations presented lower errors. The best hybrid algorithms identified in this experiment are used in the next phase.

\subsection{The second experiment}
\par At this phase, instances of up to 280 cities were the focus of the study. Results of the family KroX100 are included. The best hybridizations identified in the previous section were used to solve the selected instances. 

\par Table \ref{tab:ResultadosObtidosParaOsProblemas} presents the reference length, $Best_{ref}$, the best solution found before uncrossing, $Best_{BU}$, and after uncrossing, $Best_{AU}$, obtained solving the instances studied, the hybridization algorithm that reaches $Best_{AU}$, the OFC needed, and the stopping criterium used for each algorithm applied in this study. The error of each solution is also presented. With the proposed hybrid algorithm discussed in this paper, using general metaheuristics for discrete problems, the solutions obtained present errors equivalent to those published in the literature. For example, \cite{2017Wang} uses a hybridization approach close to the one proposed here, the errors ranging from 0\% to 4\%. \cite{2018Hussain} shows errors ranging from 0.24\% to 6.35\%. In an extreme case, the errors presented in \cite{2017Wu} vary between 2\% and 862\%. In all the mentioned cases, a specialized metaheuristic is used.
\begin{landscape}
	\begin{table}[h]
		\centering
		\caption{Results obtained with the application of the proposed approach in several instances of the symmetric TSP.} 
		\begin{tabular}{rccccccccr}
			\hline \hline
			Instance   & $N^o$ of & $Best_{ref}$ & $Best_{BU}$  & $Error_{BU}$ &  $Best_{AU}$  & $Error_{AU}$ & Hybridization        &   CFO   & Stopping    \\
			&  cities &               &              &     (\%)     &               &     (\%)     &                      &         & criterion   \\
			\hline\hline
			&        &                &              &              &               &              &  GSA+SISR2tours       & 141905 &             \\
			Eil51      &   51   &      426       &    437       &    2.58      &     432       &    1.41      &  GSA+SISR3tours       & 136241 &  300000     \\
			&        &                &              &              &               &              &  VS+SISR3tours        & 178625 &             \\\hline
			Berlin52   &   52   &      7542      &    7547      &    0.07      &     7542      &    0.00      & VS+SISR3OPT1Por       & 173791 &  300000     \\\hline
			Wg59       &   59   &      1000      &    1062      &    6.20      &     1025      &    2.50      & VS+SISR3OPT1Por       & 173791 &  300000     \\\hline
			Silalahi68 &   68   &      669       &    698       &    4.33      &     675       &    0.90      &  VS+SISR3tours        & 336724 &  500000     \\\hline
			St70       &   70   &      675       &    704       &    4.30      &     689       &    2.07      &  VS+SISR2tours        & 308296 &  500000     \\
			&        &                &              &              &               &              &  VS+SISR4tours        & 325859 &  500000     \\\hline
			Pr76       &   76   &      108159    &    115207    &    6.52      &     112741    &    4.23      &  VS+SISR3tours        & 342878 &  500000     \\\hline
			Gr96       &   96   &      55209     &    56540     &    2.41      &     56217     &    1.83      &  VS+SISR4tours        & 351623 &  500000     \\\hline
			Rat99      &   99   &      1211      &    1240      &    2.39      &     1240      &    2.39      &  VS+SISR3tours        & 346518 &  500000     \\\hline
			KroA100    &  100   &      21282     &    22169     &    4.17      &     21973     &    3.25      &  VS+SISR3OPT1Por      & 369093 &  500000     \\\hline
			KroB100    &  100   &      22141     &    23160     &    4.60      &     22647     &    2.29      &  VS+SISR4tours        & 372689 &  500000     \\\hline
			KroC100    &  100   &      20749     &    21733     &    4.74      &     21145     &    1.91      &  EA+SISR3tours        & 324840 &  500000     \\\hline
			KroD100    &  100   &      21294     &    22673     &    6.48      &     22217     &    4.33      &  GSA+SISR3tours       & 255420 &  500000     \\\hline
			KroE100    &  100   &      22068     &    23119     &    4.76      &     22495     &    1.93      &  EA+Uncross50Prob5Por & 100848 &  500000     \\\hline
			Eil101     &  101   &      629       &    669       &    6.36      &     654       &    3.97      &  VS+SISR-4tours       & 379206 &  500000     \\\hline
			Pr107      &  107   &      44303     &    44776     &    1.07      &     44510     &    0.47      &  VS+SISR3OPT1Por      & 355257 &  500000     \\\hline
			Bier127    &  127   &      118282    &    120852    &    2.17      &     120507    &    1.88      &  VS+SISR3tours        & 743430 &  1000000    \\\hline
			Ch130      &  130   &      6110      &    6392      &    4.62      &     6331      &    3.62      &  VS+SISR3tours        & 794216 &  1000000    \\\hline
			Gr137      &  137   &      69853     &    73742     &    5.57      &     70740     &    1.27      & VS+Uncross50Prob5Por  & 684417 &  1000000    \\\hline
			Ch150      &  150   &      6528      &    6828      &    4.59      &     6655      &    1.94      &  VS+SISR4tours        & 420796 &  500000     \\\hline
			KroA150    &  150   &      26524     &    29136     &    9.85      &     27903     &    5.20      &  VS+SISR2tours        & 428781 &  500000     \\\hline
			KroB150    &  150   &      26130     &    28296     &    8.29      &     27186     &    4.04      &  VS+SISR3tours        & 411380 &  500000     \\\hline
			U159       &  159   &      42080     &    43381     &    3.09      &     42749     &    1.59      &  VS+SISR2tours        & 721979 &  1000000     \\\hline
			D198       &  198   &      15780     &    16618     &    5.31      &     16158     &    2.40      &  VS+SISR4tours        & 789087 &  1000000    \\\hline
			KroA200    &  200   &      29368     &    31857     &    8.48      &     31294     &    6.56      &  VS+SISR4tours        & 843956 &  1000000    \\\hline
			KroB200    &  200   &      29437     &    32648     &   10.91      &     31099     &    5.52      &  VS+SISR3tours        & 870057 &  1000000    \\\hline
			Ts225      &  225   &      126643    &    1300725   &    3.22      &     128003    &    1.07      &  VS+SISR3tours        & 872770 &  1000000    \\\hline
			Tsp225     &  225   &      3916      &    4216      &    7.66      &     4060      &    3.68      &  VS+SISR4tours        & 883309 &  1000000    \\\hline
			A280       &  280   &      2579      &    2784      &    7.95      &     2680      &    3.92      &  VS+SISR2tours        & 873438 &  1000000    \\\hline
			\hline \hline
		\end{tabular}
		\label{tab:ResultadosObtidosParaOsProblemas}
	\end{table}
\end{landscape}

\par Only four hybridization excel in obtaining the best solutions: GSA/SISRXtour (X=2,3), VS/SISRXtours (X=2,3,4), VS/SISR3OPTYRep (Y=1\%), VS/UncrossXProbYRep (X=50\%, Y=5\%). Errors inferior to 2\% were achieved in ten instances of varied sizes. Errors superior to 4\% were obtained for only six cases.

\par The final loop of the uncrossing algorithm reduced the error of the solution by 47\% on average. In fifteen cases (more than 50\% of the cases studied), the reduction is greater than 50\%.

\par
\begin{figure}[H]
	\includegraphics[width=\linewidth]{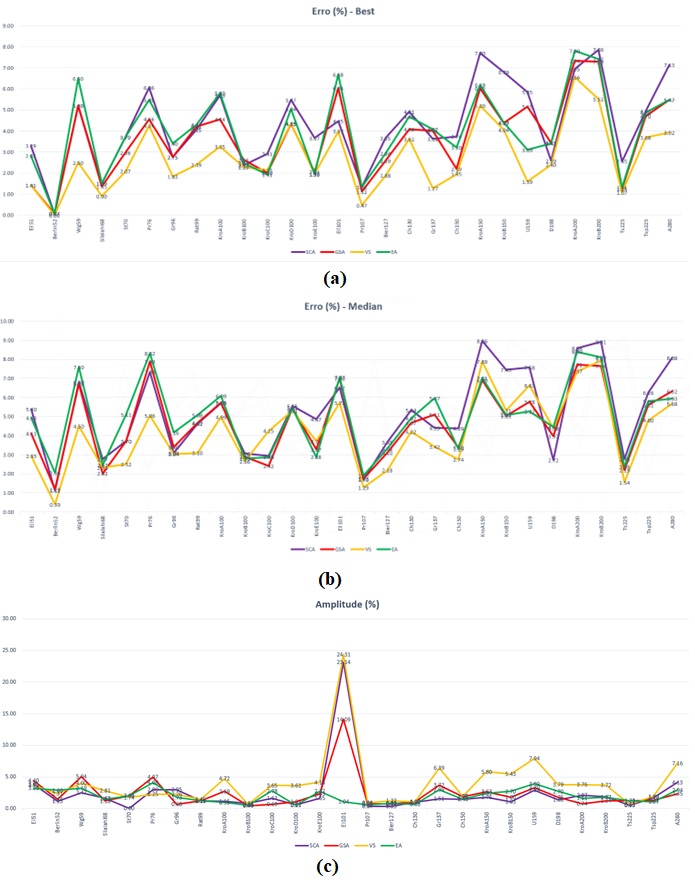}
	\caption{Average errors and amplitude of the solutions of the thirty runs for each solved instance obtained with the best hybridizations identified in this study: (a) error of the best solution, (b) error of the median, and (c) amplitude of the obtained solutions.}
	\label{fig11:ErrosTodos}
\end{figure}

 Figure \ref{fig11:ErrosTodos} shows the lower error obtained by each metaheuristic for each problem studied, no matter the hybridization. The best hybridization for each instance in Figure \ref{fig11:ErrosTodos}.a is registered in Table \ref{tab:ResultadosObtidosParaOsProblemas}. The errors are less than ten per cent. Interestingly, almost all metaheuristics hybridized converge to solutions with almost the same error for some instances, namely Berlin52, Silalahi68, KroB100, Pr107, and Ts225, but present distinct errors for other cases. The VS usually gives the best solution. The hybridization with the SISR with 2, 3 and 4 tours usually has more success. However, the hybridizations with SISR3Opt and UNCROSSXProbYRep have also obtained the best solution for some instances. The iterative uncrossing hybridization (UNCROSSXProbYRep) was particularly effective in solving the Gr137 instance, but the algorithm does not stand out for the solution of the other problems. SISR3OPT1Rep seems to be more suitable for smaller instances. 

\par Figure \ref{fig11:ErrosTodos}.b gives the error of the median of the best results found by the different hybridizations with each metaheuristic. The pattern is similar to the one presented by the best solution for each metaheuristic. Still, we can notice a spreading of the medians for some cases, such as Berlin82, KroC100, and D198, and a gathering for other instances: Gr96, Pr107, KroB150, and Ts225. VS also shows the best performance when the median is considered, which means that several hybridizations present solutions close to the best one. 

\par Generally, the amplitude of the best solutions for each hybridization, computed as the difference between the maximum and the minimum of the best solutions divided by the best one, is less than 8 per cent, no matter the metaheuristic, Figure \ref{fig11:ErrosTodos}.c. One result is particularly interesting, Eil101. The distribution of the best result of the hybridizations presents a high amplitude for the referred instance, far superior to the other cases but for hybridizations with the EA metaheuristic. However, the error of the best solution and the error associated with the median are of the same magnitude as the other instances. Finally, it is interesting to note that VS shows the greater solution amplitude in the majority of the instances studied.

\par Additional studies with more cities should be done with the SISR3OPT1Rep and UNCROSSXProbYRep hybrid algorithms to evaluate their efficacy better. Data obtained from BH and MVS do not give additional information and have been omitted.

\par
\begin{figure}[H]
	\includegraphics[width=\linewidth]{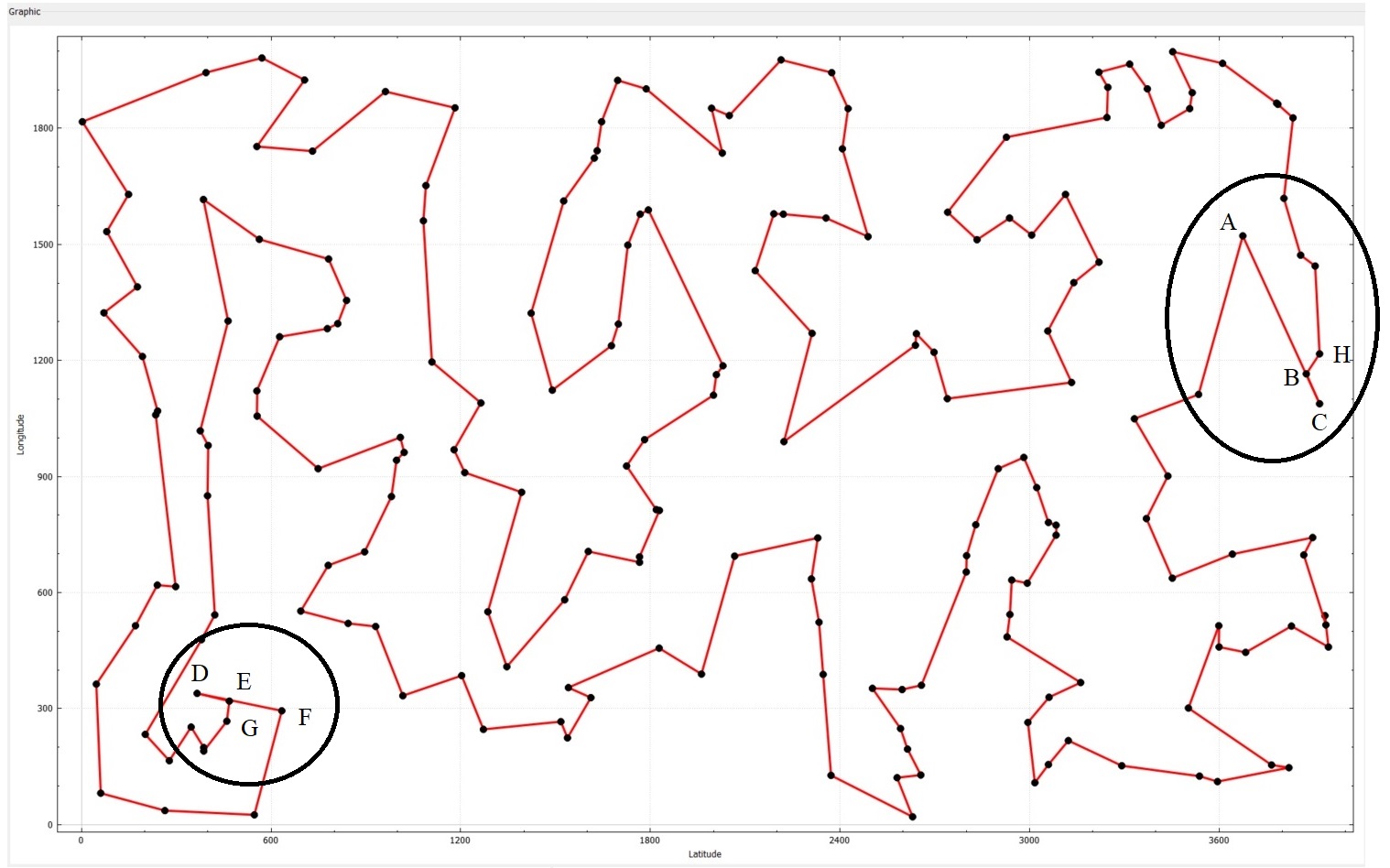}
	\caption{Illustration of the appearance of nearly parallel edges in the route corrected by the uncrossing heuristic for the KroB200 problem.}
	\label{fig12:KroB200}
\end{figure}

 It is important to note that the uncrossing heuristic used in this study does not verify two edges almost collinear, with small variation of slope. For example, the best route obtained after applying the uncrossing heuristic for the KroB200 problem is presented in Figure \ref{fig12:KroB200}.

\par Note two sets of highlighted points: the first formed by the points A = (3675, 1522), B = (3876, 1165), C = (3918, 1088) and H = (3918, 1217), and the second set formed by the points D = (3675, 1522), E = (468, 319), F = (634, 294) and G = (460, 267). It is easy to prove mathematically that these points are not collinear. In the first group taking a new sequence H-C-B-A, would have a new graph with a length smaller than the current one. The same should happen if the new sequence were F-E-D-G for the second group of points. In this way, an analysis of this situation, where there is a slope of an edge that differs slightly from a slope of other edges, can make an uncrossing heuristic more efficient, but increasing the code complexity. To exemplify the difference, changing the sequence of points for the two identified cases, the length of the route changes from 31099 to 31035, a reduction of only 0.2\%. This type of improvement of the crossover heuristic that brings gains in results for the length of the path length may be focus of future studies.

\section{Conclusion}\label{sec13}
\par The hybridization of algorithms is used to help and improve optimization techniques. As metaheuristics are very flexible optimization algorithms capable of solving different types of problems, their use has been growing in solving discrete problems such as the TSP. In general, metaheuristics deal with discrete problems using, for instance, specialised operators, such as permutation operators, to generate feasible solutions. 

\par This paper solved the classic symmetric TSP using high-level hybridization of basic metaheuristics and heuristics based on SISR, 3OPT, and line segment overlay. No change to the basic movement operators is made. Each solution randomly generated by one of the metaheuristics was corrected to become feasible with a small impact regarding the total time searching for the optimal solution. Problems with a number of cities ranging from 50 to 280 were solved. The quality of the solutions, in terms of the error regarding optimum published results, is inferior to the obtained by some low-level hybridizations and  comparable to results of other high-level hybridizations presented in the literature. The results allow one to compare the relative performance of various basic metaheuristics. 

\par Generally, the best results for 69\% of the problems were obtained with a high-level hybrid algorithm combining the SISR-like heuristic and the Vortex Search metaheuristic. However, in 31\% of the problems, good solutions were found using other hybridizations proposed in this work. Still, only 14\% were obtained with a different metaheuristic, namely the gravitational search algorithm (GSA) and the evolutionary algorithm (EA).

\par Such results suggest that the best strategy to adopt is solving the TSP using more than one hybrid algorithm with a more restricted number of runs for each hybridization. In this paper, each instance was solved thirty times, but considering the forty-two options of hybridization considered in this work, each problem was solved 1260 times. The relatively small error of the median of the best solutions found by a given metaheuristic with each hybridization reinforces this strategy, suggesting that the number of repetitions can be reduced.

\par The final loop of the uncrossing algorithm was essential to reduce the error of the best solutions from 4.9\% to 2.6\%, on average. The results also suggest that the error increases with the problem size, but additional cases should be solved to confirm this tendency.

\par Despite the relative success shown by the high-level hybridizations presented, some aspects can be improved. The heuristics were executed in cascade. However, some operations could be optimized by merging them. It is also necessary to adapt the uncrossing heuristic to treat instances with geographic coordinates. Even if the case of edges that do not cross but have almost the same linear coefficient result in a small reduction of the total path length, as illustrated in the text, the algorithm should treat them. This last case is relatively simple to identify visually but imposes some caution regarding the algorithm.

\bmhead{Acknowledgments}

This work was supported in part by the Conselho Nacional de Desenvolvimento Cient\'ifico e Tecnol\'ogico - Brasil (CNPq) under Grant $307691/2020-9$, and by the Coordena\c{c}\~ao de Aperfei\c{c}oamento de Pessoal de N\'ivel Superior - Brasil (CAPES) - Finance Code 001.

\section*{Declarations}
\subsubsection*{Conflict of Interest Statement}
Partial financial support was received from Conselho Nacional de Desenvolvimento Cient\'ifico e Tecnol\'ogico - Brasil (CNPq) under Grant $307691/2020-9$, and by the Coordena\c{c}\~ao de Aperfei\c{c}oamento de Pessoal de N\'ivel Superior - Brasil (CAPES) - Finance Code 001.

\subsubsection*{Data Availability} 
The data generated during and/or analysed during the current study are available in the TSPLIB95 (http://comopt.ifi.uni-heidelberg.de/software/TSPLIB95/), NSERC (http://www.math.uwaterloo.ca/tsp/index.html), TSP Burkardt (https://people.sc.fsu.edu/\~jburkardt/datasets/cities/cities.html) repositories, and paper (https://doi.org/10.31764/jtam.v6i3.8481).

\subsubsection*{Authors' Contributions}
Conceptualization: Carlos Alberto da Silva Junior; Data curation: Carlos Alberto da Silva Junior and Luiz Carlos Farias da Silva; Formal analysis and investigation: Carlos Alberto da Silva Junior an Angelo Passaro; Funding acquisition: Angelo Passaro; Investigation: Carlos Alberto da Silva Junior; Methodology: Carlos Alberto da Silva Junior and Angelo Passaro; Project administration: Angelo Passaro; Resources: Angelo Passaro; Software: Carlos Alberto da Silva Junior, Roberto Yuji Tanaka, and Angelo Passaro; Supervision: Angelo Passaro; Validation: Carlos Alberto da Silva Junior; Visualization: Carlos Alberto da Silva Junior and Angelo Passaro; Writing - original draft: Carlos Alberto da Silva Junior and Angelo Angelo Passaro; Writing - review \& editing: Carlos Alberto da Silva Junior, Roberto Yuji Tanaka and Angelo Passaro. All authors read and approved the final manuscript.

\bibliography{References}


\begin{thebibliography}{49}
\ifx \bisbn   \undefined \def \bisbn  #1{ISBN #1}\fi
\ifx \binits  \undefined \def \binits#1{#1}\fi
\ifx \bauthor  \undefined \def \bauthor#1{#1}\fi
\ifx \batitle  \undefined \def \batitle#1{#1}\fi
\ifx \bjtitle  \undefined \def \bjtitle#1{#1}\fi
\ifx \bvolume  \undefined \def \bvolume#1{\textbf{#1}}\fi
\ifx \byear  \undefined \def \byear#1{#1}\fi
\ifx \bissue  \undefined \def \bissue#1{#1}\fi
\ifx \bfpage  \undefined \def \bfpage#1{#1}\fi
\ifx \blpage  \undefined \def \blpage #1{#1}\fi
\ifx \burl  \undefined \def \burl#1{\textsf{#1}}\fi
\ifx \doiurl  \undefined \def \doiurl#1{\url{https://doi.org/#1}}\fi
\ifx \betal  \undefined \def \betal{\textit{et al.}}\fi
\ifx \binstitute  \undefined \def \binstitute#1{#1}\fi
\ifx \binstitutionaled  \undefined \def \binstitutionaled#1{#1}\fi
\ifx \bctitle  \undefined \def \bctitle#1{#1}\fi
\ifx \beditor  \undefined \def \beditor#1{#1}\fi
\ifx \bpublisher  \undefined \def \bpublisher#1{#1}\fi
\ifx \bbtitle  \undefined \def \bbtitle#1{#1}\fi
\ifx \bedition  \undefined \def \bedition#1{#1}\fi
\ifx \bseriesno  \undefined \def \bseriesno#1{#1}\fi
\ifx \blocation  \undefined \def \blocation#1{#1}\fi
\ifx \bsertitle  \undefined \def \bsertitle#1{#1}\fi
\ifx \bsnm \undefined \def \bsnm#1{#1}\fi
\ifx \bsuffix \undefined \def \bsuffix#1{#1}\fi
\ifx \bparticle \undefined \def \bparticle#1{#1}\fi
\ifx \barticle \undefined \def \barticle#1{#1}\fi
\bibcommenthead
\ifx \bconfdate \undefined \def \bconfdate #1{#1}\fi
\ifx \botherref \undefined \def \botherref #1{#1}\fi
\ifx \url \undefined \def \url#1{\textsf{#1}}\fi
\ifx \bchapter \undefined \def \bchapter#1{#1}\fi
\ifx \bbook \undefined \def \bbook#1{#1}\fi
\ifx \bcomment \undefined \def \bcomment#1{#1}\fi
\ifx \oauthor \undefined \def \oauthor#1{#1}\fi
\ifx \citeauthoryear \undefined \def \citeauthoryear#1{#1}\fi
\ifx \endbibitem  \undefined \def \endbibitem {}\fi
\ifx \bconflocation  \undefined \def \bconflocation#1{#1}\fi
\ifx \arxivurl  \undefined \def \arxivurl#1{\textsf{#1}}\fi
\csname PreBibitemsHook\endcsname

\bibitem[\protect\citeauthoryear{Saji and Riffi}{2015}]{2015Saji}
\begin{barticle}
\bauthor{\bsnm{Saji}, \binits{Y.}},
\bauthor{\bsnm{Riffi}, \binits{M.E.}}:
\batitle{A novel discrete bat algorithm for solving the travelling salesman
  problem}.
\bjtitle{Neural computing \& applications}
\bvolume{27}(\bissue{7}),
\bfpage{1853}--\blpage{1866}
(\byear{2015})
\end{barticle}
\endbibitem

\bibitem[\protect\citeauthoryear{Frieze et~al.}{1982}]{1982Frieze}
\begin{barticle}
\bauthor{\bsnm{Frieze}, \binits{A.M.}},
\bauthor{\bsnm{Galbiati}, \binits{G.}},
\bauthor{\bsnm{Maffioli}, \binits{F.}}:
\batitle{On the worst-case performance of some algorithms for the asymmetric
  traveling salesman problem}.
\bjtitle{Networks}
\bvolume{12},
\bfpage{23}--\blpage{39}
(\byear{1982})
\end{barticle}
\endbibitem

\bibitem[\protect\citeauthoryear{Li et~al.}{2015}]{2015Li}
\begin{barticle}
\bauthor{\bsnm{Li}, \binits{J.}},
\bauthor{\bsnm{Zhou}, \binits{M.M.}},
\bauthor{\bsnm{Sun}, \binits{Q.}},
\bauthor{\bsnm{Dai}, \binits{X.}},
\bauthor{\bsnm{Yu}, \binits{X.}}:
\batitle{Colored traveling salesman problem}.
\bjtitle{IEEE Transactions on Cybernetics}
\bvolume{45}(\bissue{11}),
\bfpage{2390}--\blpage{2401}
(\byear{2015})
\doiurl{10.1109/TCYB.2014.2371918}
\end{barticle}
\endbibitem

\bibitem[\protect\citeauthoryear{Chisman}{1975}]{1975Chisman}
\begin{barticle}
\bauthor{\bsnm{Chisman}, \binits{J.A.}}:
\batitle{The clustered traveling salesman problem}.
\bjtitle{Computers \& operations research}
\bvolume{2}(\bissue{2}),
\bfpage{115}--\blpage{119}
(\byear{1975})
\end{barticle}
\endbibitem

\bibitem[\protect\citeauthoryear{Laporte and Nobert}{1983}]{1983Laporte}
\begin{barticle}
\bauthor{\bsnm{Laporte}, \binits{G.}},
\bauthor{\bsnm{Nobert}, \binits{Y.}}:
\batitle{Generalized travelling salesman problem through n sets of nodes: An
  integer programming approach}.
\bjtitle{INFOR: Information Systems and Operational Research}
\bvolume{21}(\bissue{1}),
\bfpage{61}--\blpage{75}
(\byear{1983})
\doiurl{10.1080/03155986.1983.11731885}
\end{barticle}
\endbibitem

\bibitem[\protect\citeauthoryear{Powell et~al.}{1995}]{1995Powell}
\begin{bchapter}
\bauthor{\bsnm{Powell}, \binits{W.B.}},
\bauthor{\bsnm{Jaillet}, \binits{P.}},
\bauthor{\bsnm{Odoni}, \binits{A.}}:
\bctitle{Chapter 3 stochastic and dynamic networks and routing}.
In: \beditor{\bsnm{xxxxxxxx}} (ed.)
\bbtitle{Network Routing},
pp. \bfpage{141}--\blpage{295}.
\bpublisher{Elsevier}, \blocation{???}
(\byear{1995})
\end{bchapter}
\endbibitem

\bibitem[\protect\citeauthoryear{Obermeyer et~al.}{2012}]{2012Obermeyer}
\begin{barticle}
\bauthor{\bsnm{Obermeyer}, \binits{K.J.}},
\bauthor{\bsnm{Oberlin}, \binits{P.}},
\bauthor{\bsnm{Darbha}, \binits{S.}}:
\batitle{Sampling-based path planning for a visual reconnaissance unmanned air
  vehicle}.
\bjtitle{Journal of Guidance, Control, and Dynamics}
\bvolume{35}(\bissue{2}),
\bfpage{619}--\blpage{631}
(\byear{2012})
\doiurl{10.2514/1.48949}
\end{barticle}
\endbibitem

\bibitem[\protect\citeauthoryear{Roberti and Ruthmair}{2021}]{2021Roberti}
\begin{barticle}
\bauthor{\bsnm{Roberti}, \binits{R.}},
\bauthor{\bsnm{Ruthmair}, \binits{M.}}:
\batitle{Exact methods for the traveling salesman problem with drone}.
\bjtitle{Transportation Science}
\bvolume{55}(\bissue{2}),
\bfpage{315}--\blpage{335}
(\byear{2021})
\doiurl{10.1287/trsc.2020.1017}
\end{barticle}
\endbibitem

\bibitem[\protect\citeauthoryear{Battarra et~al.}{2010}]{2010Battarra}
\begin{barticle}
\bauthor{\bsnm{Battarra}, \binits{M.}},
\bauthor{\bsnm{Erdoǧan}, \binits{G.}},
\bauthor{\bsnm{Laporte}, \binits{G.}},
\bauthor{\bsnm{Vigo}, \binits{D.}}:
\batitle{The traveling salesman problem with pickups, deliveries, and handling
  costs}.
\bjtitle{Transportation Science}
\bvolume{44}(\bissue{3}),
\bfpage{383}--\blpage{399}
(\byear{2010})
\end{barticle}
\endbibitem

\bibitem[\protect\citeauthoryear{Savelsbergh}{1985}]{1985Savelsbergh}
\begin{barticle}
\bauthor{\bsnm{Savelsbergh}, \binits{M.W.P.}}:
\batitle{Local search in routing problems with time windows}.
\bjtitle{Annals of Operations Research}
\bvolume{4}(\bissue{1}),
\bfpage{285}--\blpage{305}
(\byear{1985})
\doiurl{10.1007/BF02022044}
\end{barticle}
\endbibitem

\bibitem[\protect\citeauthoryear{{De Lima Filho}
  et~al.}{2022}]{2022DeLimaFilho}
\begin{barticle}
\bauthor{\bsnm{{De Lima Filho}}, \binits{G.M.}},
\bauthor{\bsnm{Passaro}, \binits{A.}},
\bauthor{\bsnm{Delfino}, \binits{G.M.}},
\bauthor{\bsnm{Santana}, \binits{L.D.}},
\bauthor{\bsnm{Monsuur}, \binits{H.}}:
\batitle{Time-critical maritime uav mission planning using a neural network: An
  operational view}.
\bjtitle{IEEE Access}
\bvolume{10},
\bfpage{111749}--\blpage{111758}
(\byear{2022})
\doiurl{10.1109/ACCESS.2022.3215646}
\end{barticle}
\endbibitem

\bibitem[\protect\citeauthoryear{Christofides}{1985}]{1985Christofides}
\begin{bbook}
\bauthor{\bsnm{Christofides}, \binits{N.}}:
In: \beditor{\bsnm{Lawler}, \binits{E.L.}},
\beditor{\bsnm{Lenstra}, \binits{J.K.}},
\beditor{\bsnm{Kan}, \binits{A.H.G.R.}},
\beditor{\bsnm{Shmoys}, \binits{D.B.}} (eds.)
\bbtitle{The Vehicle Routing Problem}.
\bpublisher{Wiley}, \blocation{???}
(\byear{1985})
\end{bbook}
\endbibitem

\bibitem[\protect\citeauthoryear{Dantzig et~al.}{1954}]{1954Dantzig}
\begin{barticle}
\bauthor{\bsnm{Dantzig}, \binits{G.}},
\bauthor{\bsnm{Fulkerson}, \binits{R.}},
\bauthor{\bsnm{Johnson}, \binits{S.}}:
\batitle{It is shown that a certain tour of 49 cities, one in each of the 48
  states and washington, d. c., has the shortest road distance}.
\bjtitle{Journal of the Operations Research Society of America}
\bvolume{2}(\bissue{4}),
\bfpage{393}--\blpage{410}
(\byear{1954})
\end{barticle}
\endbibitem

\bibitem[\protect\citeauthoryear{Held and Karp}{1970}]{1970Held}
\begin{barticle}
\bauthor{\bsnm{Held}, \binits{M.}},
\bauthor{\bsnm{Karp}, \binits{R.M.}}:
\batitle{The traveling-salesman problem and minimum spanning trees.}
\bjtitle{Operations Research}
\bvolume{18}(\bissue{6}),
\bfpage{1138}--\blpage{1162}
(\byear{1970})
\doiurl{http://www.jstor.org/stable/169411}
\end{barticle}
\endbibitem

\bibitem[\protect\citeauthoryear{Lin and Kernighan}{1973}]{1973Lin}
\begin{barticle}
\bauthor{\bsnm{Lin}, \binits{S.}},
\bauthor{\bsnm{Kernighan}, \binits{B.W.}}:
\batitle{An effective heuristic algorithm for the traveling-salesman problem}.
\bjtitle{Operations Research}
\bvolume{21}(\bissue{2}),
\bfpage{498}--\blpage{516}
(\byear{1973})
\end{barticle}
\endbibitem

\bibitem[\protect\citeauthoryear{Hopfield and Tank}{1985}]{1985Hopfield}
\begin{barticle}
\bauthor{\bsnm{Hopfield}, \binits{J.J.}},
\bauthor{\bsnm{Tank}, \binits{D.W.}}:
\batitle{“neural” computation of decisions in optimization problems}.
\bjtitle{Biol. Cybern}
\bvolume{52},
\bfpage{141}--\blpage{152}
(\byear{1985})
\doiurl{10.1007/BF00339943}
\end{barticle}
\endbibitem

\bibitem[\protect\citeauthoryear{Arigliano et~al.}{2019}]{2019Arigliano}
\begin{barticle}
\bauthor{\bsnm{Arigliano}, \binits{A.}},
\bauthor{\bsnm{Ghiani}, \binits{G.}},
\bauthor{\bsnm{Grieco}, \binits{A.}},
\bauthor{\bsnm{Guerriero}, \binits{E.}},
\bauthor{\bsnm{Plana}, \binits{I.}}:
\batitle{Time-dependent asymmetric traveling salesman problem with time
  windows: Properties and an exact algorithm}.
\bjtitle{Discrete Applied Mathematics}
\bvolume{261},
\bfpage{28}--\blpage{39}
(\byear{2019})
\doiurl{10.1016/j.dam.2018.09.017}
\end{barticle}
\endbibitem

\bibitem[\protect\citeauthoryear{Dong and Cai}{2019}]{2019Dong}
\begin{barticle}
\bauthor{\bsnm{Dong}, \binits{X.}},
\bauthor{\bsnm{Cai}, \binits{Y.}}:
\batitle{A novel genetic algorithm for large scale colored balanced traveling
  salesman problem}.
\bjtitle{Future Generation Computer Systems}
\bvolume{95},
\bfpage{727}--\blpage{742}
(\byear{2019})
\doiurl{10.1016/j.future.2018.12.065}
\end{barticle}
\endbibitem

\bibitem[\protect\citeauthoryear{Osaba et~al.}{2018}]{2018Osaba}
\begin{bbook}
\bauthor{\bsnm{Osaba}, \binits{E.}},
\bauthor{\bsnm{{Del Ser}}, \binits{J.}},
\bauthor{\bsnm{Iglesias}, \binits{A.}},
\bauthor{\bsnm{Bilbao}, \binits{N.}},
\bauthor{\bsnm{Fister}, \binits{I.}},
\bauthor{\bsnm{{Fister Jr}}, \binits{I.}},
\bauthor{\bsnm{Galvez}, \binits{A.}}:
\bbtitle{Solving the Open-Path Asymmetric Green Traveling Salesman Problem in a
  Realistic Urban Environment},
pp. \bfpage{181}--\blpage{191}
(\byear{2018})
\end{bbook}
\endbibitem

\bibitem[\protect\citeauthoryear{Ezugwu et~al.}{2017}]{2017Ezugwu}
\begin{barticle}
\bauthor{\bsnm{Ezugwu}, \binits{A.E.-S.}},
\bauthor{\bsnm{Ezugwu}, \binits{A.O.}},
\bauthor{\bsnm{Frîncu}, \binits{M.E.}}:
\batitle{Simulated annealing based symbiotic organisms search optimization
  algorithm for traveling salesman problem}.
\bjtitle{Expert Systems with Applications}
\bvolume{77},
\bfpage{189}--\blpage{210}
(\byear{2017})
\doiurl{10.1016/j.eswa.2017.01.053}
\end{barticle}
\endbibitem

\bibitem[\protect\citeauthoryear{Hussain et~al.}{2019}]{2018Hussain}
\begin{barticle}
\bauthor{\bsnm{Hussain}, \binits{A.}},
\bauthor{\bsnm{Muhammad}, \binits{Y.S.}},
\bauthor{\bsnm{Sajid}, \binits{N.}}:
\batitle{A simulated study of genetic algorithm with a new crossover operator
  using traveling salesman problem}.
\bjtitle{Journal of Mathematics}
\bvolume{51}(\bissue{5}),
\bfpage{61}--\blpage{77}
(\byear{2019})
\end{barticle}
\endbibitem

\bibitem[\protect\citeauthoryear{Zhong et~al.}{2018}]{2018Zhong}
\begin{barticle}
\bauthor{\bsnm{Zhong}, \binits{Y.}},
\bauthor{\bsnm{Lin}, \binits{J.}},
\bauthor{\bsnm{Wang}, \binits{L.}},
\bauthor{\bsnm{Zhang}, \binits{H.}}:
\batitle{Discrete comprehensive learning particle swarm optimization algorithm
  with metropolis acceptance criterion for traveling salesman problem}.
\bjtitle{Swarm and Evolutionary Computation}
\bvolume{42},
\bfpage{77}--\blpage{88}
(\byear{2018})
\doiurl{10.1016/j.swevo.2018.02.017}
\end{barticle}
\endbibitem

\bibitem[\protect\citeauthoryear{Liang et~al.}{2006}]{2006Liang}
\begin{barticle}
\bauthor{\bsnm{Liang}, \binits{J.J.}},
\bauthor{\bsnm{Qin}, \binits{A.K.}},
\bauthor{\bsnm{Suganthan}, \binits{P.N.}},
\bauthor{\bsnm{Baskar}, \binits{S.}}:
\batitle{Comprehensive learning particle swarm optimizer for global
  optimization of multimodal functions}.
\bjtitle{IEEE Transactions on Evolutionary Computation}
\bvolume{10}(\bissue{3}),
\bfpage{281}--\blpage{295}
(\byear{2006})
\doiurl{10.1109/TEVC.2005.857610}
\end{barticle}
\endbibitem

\bibitem[\protect\citeauthoryear{Chen et~al.}{2010}]{2010Chen}
\begin{barticle}
\bauthor{\bsnm{Chen}, \binits{W.N.}},
\bauthor{\bsnm{Zhang}, \binits{J.}},
\bauthor{\bsnm{Chung}, \binits{H.S.H.}},
\bauthor{\bsnm{Zhong}, \binits{W.L.}},
\bauthor{\bsnm{Wu}, \binits{W.G.}},
\bauthor{\bsnm{Shi}, \binits{Y.H.}}:
\batitle{A novel set-based particle swarm optimization method for discrete
  optimization problems}.
\bjtitle{IEEE Transactions on Evolutionary Computation}
\bvolume{14}(\bissue{2}),
\bfpage{278}--\blpage{300}
(\byear{2010})
\doiurl{10.1109/TEVC.2009.2030331}
\end{barticle}
\endbibitem

\bibitem[\protect\citeauthoryear{Osaba et~al.}{2020}]{2020Osaba}
\begin{bchapter}
\bauthor{\bsnm{Osaba}, \binits{E.}},
\bauthor{\bsnm{Yang}, \binits{X.S.}},
\bauthor{\bsnm{{Del Ser}}, \binits{J.}}:
\bctitle{Chapter 9 - traveling salesman problem: a perspective review of recent
  research and new results with bio-inspired metaheuristics}.
In: \beditor{\bsnm{Yang}, \binits{X.S.}} (ed.)
\bbtitle{Nature-Inspired Computation and Swarm Intelligence},
pp. \bfpage{135}--\blpage{164}.
\bpublisher{Academic Press}, \blocation{???}
(\byear{2020}).
\doiurl{10.1016/B978-0-12-819714-1.00020-8}
\end{bchapter}
\endbibitem

\bibitem[\protect\citeauthoryear{Lehman and Stanley}{2008}]{2008Lehman}
\begin{bchapter}
\bauthor{\bsnm{Lehman}, \binits{J.}},
\bauthor{\bsnm{Stanley}, \binits{K.O.}}:
\bctitle{Exploiting open-endedness to solve problems through the search for
  novelty}.
In: \beditor{\bsnm{Bullock}, \binits{S.}},
\beditor{\bsnm{Noble}, \binits{J.}},
\beditor{\bsnm{Watson}, \binits{R.}},
\beditor{\bsnm{Bedau}, \binits{M.A.}} (eds.)
\bbtitle{Proceedings of the Eleventh International Conference on Artificial
  Life (ALIFE XI)},
pp. \bfpage{250}--\blpage{257}.
\bpublisher{MIT Press}, \blocation{???}
(\byear{2008})
\end{bchapter}
\endbibitem

\bibitem[\protect\citeauthoryear{Wu and Gao}{2017}]{2017Wu}
\begin{bchapter}
\bauthor{\bsnm{Wu}, \binits{X.}},
\bauthor{\bsnm{Gao}, \binits{D.}}:
\bctitle{A study on greedy search to improve simulated annealing for
  large-scale traveling salesman problem}.
In: \beditor{\bsnm{Tan}, \binits{Y.}},
\beditor{\bsnm{Takagi}, \binits{H.}},
\beditor{\bsnm{Shi}, \binits{Y.}},
\beditor{\bsnm{Niu}, \binits{B.}} (eds.)
\bbtitle{Advances in Swarm Intelligence},
pp. \bfpage{250}--\blpage{257}.
\bpublisher{Springer}, \blocation{???}
(\byear{2017})
\end{bchapter}
\endbibitem

\bibitem[\protect\citeauthoryear{Eskandari et~al.}{2019}]{2019Eskandari}
\begin{bbook}
\bauthor{\bsnm{Eskandari}, \binits{L.}},
\bauthor{\bsnm{Jafarian}, \binits{A.}},
\bauthor{\bsnm{Rahimloo}, \binits{P.}},
\bauthor{\bsnm{Baleanu}, \binits{D.}}:
In: \beditor{\bsnm{Ta{\c{s}}}, \binits{K.}},
\beditor{\bsnm{Baleanu}, \binits{D.}},
\beditor{\bsnm{Machado}, \binits{J.A.T.}} (eds.)
\bbtitle{A Modified and Enhanced Ant Colony Optimization Algorithm for
  Traveling Salesman Problem},
pp. \bfpage{257}--\blpage{265}.
\bpublisher{Springer}, \blocation{???}
(\byear{2019})
\end{bbook}
\endbibitem

\bibitem[\protect\citeauthoryear{Wang and Xu}{2017}]{2017Wang}
\begin{barticle}
\bauthor{\bsnm{Wang}, \binits{Y.}},
\bauthor{\bsnm{Xu}, \binits{N.}}:
\batitle{A hybrid particle swarm optimization method for traveling salesman
  problem}.
\bjtitle{International Journal of Applied Metaheuristic Computing (IJAMC)}
\bvolume{8},
\bfpage{53}--\blpage{65}
(\byear{2017})
\doiurl{10.4018/IJAMC.2017070104}
\end{barticle}
\endbibitem

\bibitem[\protect\citeauthoryear{Flood}{1956}]{1956Flood}
\begin{botherref}
\oauthor{\bsnm{Flood}, \binits{M.M.}}:
The traveling-salesman problem.
Operations Research
\textbf{4}(1)
(1956)
\end{botherref}
\endbibitem

\bibitem[\protect\citeauthoryear{Mazidi et~al.}{2016}]{2016Mazidi}
\begin{barticle}
\bauthor{\bsnm{Mazidi}, \binits{A.}},
\bauthor{\bsnm{Fakhrahmad}, \binits{M.}},
\bauthor{\bsnm{Sadreddini}, \binits{M.}}:
\batitle{A meta-heuristic approach to cvrp problem: Local search optimization
  based on ga and ant colony}.
\bjtitle{Journal of Advances in Computer Research}
\bvolume{7}(\bissue{1}),
\bfpage{1}--\blpage{22}
(\byear{2016})
\end{barticle}
\endbibitem

\bibitem[\protect\citeauthoryear{Marinakis and Marinaki}{2010}]{2010Marinakis}
\begin{barticle}
\bauthor{\bsnm{Marinakis}, \binits{Y.}},
\bauthor{\bsnm{Marinaki}, \binits{M.}}:
\batitle{A hybrid genetic - particle swarm optimization algorithm for the
  vehicle routing problem}.
\bjtitle{Expert Systems with Applications}
\bvolume{37}(\bissue{2}),
\bfpage{1446}--\blpage{1455}
(\byear{2010})
\doiurl{10.1016/j.eswa.2009.06.085}
\end{barticle}
\endbibitem

\bibitem[\protect\citeauthoryear{Christiaens and
  Berghe}{2020}]{2020Christiaens}
\begin{barticle}
\bauthor{\bsnm{Christiaens}, \binits{J.}},
\bauthor{\bsnm{Berghe}, \binits{G.V.}}:
\batitle{{Slack Induction by String Removals for Vehicle Routing Problems}}.
\bjtitle{Transportation Science}
\bvolume{54}(\bissue{2}),
\bfpage{417}--\blpage{433}
(\byear{2020})
\end{barticle}
\endbibitem

\bibitem[\protect\citeauthoryear{Talbi}{2009}]{2009Talbi}
\begin{bbook}
\bauthor{\bsnm{Talbi}, \binits{E.-G.}}:
\bbtitle{Metaheuristics - From Design to Implementation},
\bedition{1st} edn.
\bpublisher{John Wiley and Sons},
\blocation{New Jersey}
(\byear{2009})
\end{bbook}
\endbibitem

\bibitem[\protect\citeauthoryear{Kennedy and Eberhart}{1995}]{1995Kennedy}
\begin{bchapter}
\bauthor{\bsnm{Kennedy}, \binits{J.}},
\bauthor{\bsnm{Eberhart}, \binits{R.}}:
\bctitle{Particle swarm optimization}.
In: \bbtitle{Neural Networks, 1995. Proceedings., IEEE International Conference
  On},
pp. \bfpage{1942}--\blpage{1948}.
\bpublisher{IEEE},
\blocation{Perth, WA, Australia}
(\byear{1995})
\end{bchapter}
\endbibitem

\bibitem[\protect\citeauthoryear{Mirjalili}{2016}]{2016Mirjalili}
\begin{barticle}
\bauthor{\bsnm{Mirjalili}, \binits{S.}}:
\batitle{Sca: A sine cosine algorithm for solving optimization problems}.
\bjtitle{Knowledge-Based Systems}
\bvolume{96},
\bfpage{120}--\blpage{133}
(\byear{2016})
\doiurl{10.1016/j.knosys.2015.12.022}
\end{barticle}
\endbibitem

\bibitem[\protect\citeauthoryear{Dogan and Olmez}{2015}]{2015Dogan}
\begin{barticle}
\bauthor{\bsnm{Dogan}, \binits{B.}},
\bauthor{\bsnm{Olmez}, \binits{T.}}:
\batitle{Vortex search algorithm for the analog active filter component
  selection problem}.
\bjtitle{AEU - International Journal of Electronics and Communications}
\bvolume{69}(\bissue{9}),
\bfpage{1243}--\blpage{1253}
(\byear{2015})
\doiurl{10.1016/j.aeue.2015.05.005}
\end{barticle}
\endbibitem

\bibitem[\protect\citeauthoryear{Dogan}{2016}]{2016Dogan}
\begin{barticle}
\bauthor{\bsnm{Dogan}, \binits{B.}}:
\batitle{A modified vortex search algorithm for numerical function
  optimization}.
\bjtitle{International Journal of Artificial Intelligence \& Applications}
\bvolume{7}(\bissue{3}),
\bfpage{37}--\blpage{54}
(\byear{2016})
\doiurl{10.5121/ijaia.2016.7304}
\end{barticle}
\endbibitem

\bibitem[\protect\citeauthoryear{Rashedi et~al.}{2009}]{2009Rashedi}
\begin{barticle}
\bauthor{\bsnm{Rashedi}, \binits{E.}},
\bauthor{\bsnm{Nezamabadi-pour}, \binits{H.}},
\bauthor{\bsnm{Saryazdi}, \binits{S.}}:
\batitle{Gsa: A gravitational search algorithm}.
\bjtitle{Information Sciences}
\bvolume{179}(\bissue{13}),
\bfpage{2232}--\blpage{2248}
(\byear{2009})
\doiurl{10.1016/j.ins.2009.03.004}
\end{barticle}
\endbibitem

\bibitem[\protect\citeauthoryear{Kumar et~al.}{2015}]{2015Kumar}
\begin{bbook}
\bauthor{\bsnm{Kumar}, \binits{S.}},
\bauthor{\bsnm{Datta}, \binits{D.}},
\bauthor{\bsnm{Singh}, \binits{S.K.}}:
In: \beditor{\bsnm{Azar}, \binits{A.T.}},
\beditor{\bsnm{Vaidyanathan}, \binits{S.}} (eds.)
\bbtitle{Black Hole Algorithm and Its Applications},
pp. \bfpage{147}--\blpage{170}.
\bpublisher{Springer}, \blocation{???}
(\byear{2015})
\end{bbook}
\endbibitem

\bibitem[\protect\citeauthoryear{Kirkpatrick et~al.}{1983}]{1983Kirkpatrick}
\begin{barticle}
\bauthor{\bsnm{Kirkpatrick}, \binits{S.}},
\bauthor{\bsnm{Gelatt}, \binits{C.D.}},
\bauthor{\bsnm{Vecchi}, \binits{M.P.}}:
\batitle{Optimization by simulated annealing}.
\bjtitle{Science}
\bvolume{220},
\bfpage{671}--\blpage{680}
(\byear{1983})
\end{barticle}
\endbibitem

\bibitem[\protect\citeauthoryear{Schrimpf et~al.}{2000}]{2000Schrimpf}
\begin{barticle}
\bauthor{\bsnm{Schrimpf}, \binits{G.}},
\bauthor{\bsnm{Schneider}, \binits{J.}},
\bauthor{\bsnm{Wilbrandt}, \binits{H.S.}},
\bauthor{\bsnm{Dueck}, \binits{G.}}:
\batitle{Record breaking optimization results using the ruin and recreate
  principle}.
\bjtitle{Journal of Computational Physics}
\bvolume{159}(\bissue{2}),
\bfpage{139}--\blpage{171}
(\byear{2000})
\doiurl{10.1006/jcph.1999.6413}
\end{barticle}
\endbibitem

\bibitem[\protect\citeauthoryear{Silalahi et~al.}{2022}]{2022Silalahi}
\begin{barticle}
\bauthor{\bsnm{Silalahi}, \binits{B.P.}},
\bauthor{\bsnm{Sahara}, \binits{F.}},
\bauthor{\bsnm{Hanum}, \binits{F.}},
\bauthor{\bsnm{Mayyani}, \binits{H.}}:
\batitle{Simulated annealing algorithm for determining travelling salesman
  problem solution and its comparison with branch and bound method}.
\bjtitle{JTAM (Jurnal Teori dan Aplikasi Matematika)}
(\byear{2022})
\doiurl{10.31764/jtam.v6i3.8481}
\end{barticle}
\endbibitem

\bibitem[\protect\citeauthoryear{Reinelt}{1991}]{1991Reinelt}
\begin{barticle}
\bauthor{\bsnm{Reinelt}, \binits{G.}}:
\batitle{Tsplib--a traveling salesman problem library}.
\bjtitle{INFORMS journal on computing}
\bvolume{3}(\bissue{4}),
\bfpage{376}--\blpage{384}
(\byear{1991})
\end{barticle}
\endbibitem

\bibitem[\protect\citeauthoryear{Sciences and
  of~Canada~(NSERC)}{2009}]{TSPCanada}
\begin{botherref}
\oauthor{\bsnm{Sciences}, \binits{N.}},
\oauthor{\bsnm{Canada~(NSERC)}, \binits{E.R.C.}}:
Traveling Salesman Problem.
\url{http://www.math.uwaterloo.ca/tsp/index.html}
Accessed 2021-08-12
\end{botherref}
\endbibitem

\bibitem[\protect\citeauthoryear{Burkardt}{2019}]{TSPBurkardt}
\begin{botherref}
\oauthor{\bsnm{Burkardt}, \binits{J.}}:
Data for the Traveling Salesperson Problem.
\url{https://people.sc.fsu.edu/\\texttildelow
  jburkardt/datasets/cities/cities.html}
Accessed 2022-08-01
\end{botherref}
\endbibitem

\bibitem[\protect\citeauthoryear{Jiang et~al.}{2007}]{Jiang2007}
\begin{barticle}
\bauthor{\bsnm{Jiang}, \binits{M.}},
\bauthor{\bsnm{Luo}, \binits{Y.P.}},
\bauthor{\bsnm{Yang}, \binits{S.Y.}}:
\batitle{Stochastic convergence analysis and parameter selection of the
  standard particle swarm optimization algorithm}.
\bjtitle{Information Processing Letters}
\bvolume{102}(\bissue{1}),
\bfpage{8}--\blpage{16}
(\byear{2007})
\doiurl{10.1016/j.ipl.2006.10.005}
\end{barticle}
\endbibitem

\bibitem[\protect\citeauthoryear{Laarhoven and Aarts}{1987}]{Laarhoven1987}
\begin{bbook}
\beditor{\bsnm{Laarhoven}, \binits{P.J.M.}},
\beditor{\bsnm{Aarts}, \binits{E.H.L.}} (eds.):
\bbtitle{Simulated Annealing: Theory and Applications}.
\bpublisher{Kluwer Academic Publishers},
\blocation{Norwell, MA, USA}
(\byear{1987})
\end{bbook}
\endbibitem

\bibitem[\protect\citeauthoryear{TSPLIB}{1995}]{TSPLibi95}
\begin{botherref}
\oauthor{\bsnm{TSPLIB}}:
TSPLIB.
\url{http://comopt.ifi.uni-heidelberg.de/software/TSPLIB95/}
Accessed 2022-06-10
\end{botherref}
\endbibitem

\end{thebibliography}

\end{document}